\documentclass[acmsmall]{acmart}
\usepackage{ragged2e}
\usepackage{booktabs,makecell, multirow, tabularx}
\usepackage{amsmath}
\usepackage{colortbl}
\usepackage{color,xcolor}

\definecolor{CornflowerBlue}{RGB}{100,149,237} 
\definecolor{RoyalBlue}{RGB}{65,105,225}       
\definecolor{TealBlue}{RGB}{0,128,128}         
\definecolor{Green}{RGB}{0,128,0}
\definecolor{Black}{rgb}{0,0,0}

\usepackage{tikz}
\usepackage[edges]{forest}
\definecolor{hidden-draw}{RGB}{20,68,106}
\definecolor{hidden-pink}{RGB}{255,245,247}

\AtBeginDocument{%
  }

\acmJournal{JACM}

\usepackage{color}
\usepackage[most]{tcolorbox}
\definecolor{codegreen}{rgb}{0,0.6,0}
\definecolor{codegray}{rgb}{0.5,0.5,0.5}
\definecolor{codepurple}{rgb}{0.58,0,0.82}
\definecolor{backcolour}{rgb}{0.95,0.95,0.92}
\definecolor{DarkGreen}{RGB}{0,100,0}
\definecolor{DarkYellow}{rgb}{0.8, 0.8, 0.0} 
\definecolor{DarkBrown}{rgb}{0.4, 0.2, 0.1} 
\definecolor{DarkBlue}{rgb}{0.0, 0.0, 0.5} 
\definecolor{DarkRed}{rgb}{0.5, 0.0, 0.0} 

\usepackage{pifont}

\renewcommand\footnotetextcopyrightpermission[1]{}
\begin{document}

\title{Multimodal Large Language Models Meet Multimodal Emotion Recognition and Reasoning: A Survey}


\author{Yuntao~Shou}
\email{shouyuntao@stu.xjtu.edu.cn}
\affiliation{
  \institution{Central South University of Forestry and Technology}
  \city{ChangSha}
  \state{Hunan}
  \country{China}
  \postcode{410004}
}

\author{Tao Meng}
\authornote{Corresponding Author}
\email{mengtao@hnu.edu.cn}
\affiliation{
	\institution{Central South University of Forestry and Technology}
	\city{ChangSha}
	\state{Hunan}
	\country{China}
    \postcode{410004}
}

\author{Wei~Ai}
\email{aiwei@hnu.edu.cn}
\affiliation{
	\institution{Central South University of Forestry and Technology}
	\city{ChangSha}
	\state{Hunan}
	\country{China}
    \postcode{410004}
}

\author{Keqin~Li}
\email{lik@newpaltz.edu}
\affiliation{%
	\institution{State University of New York}
	\city{New Paltz}
	\state{New York}
	\country{USA}
	\postcode{12561}
}

\renewcommand{\shortauthors}{Shou et al.}

\begin{abstract}
In recent years, large language models (LLMs) have driven major advances in language understanding, marking a significant step toward artificial general intelligence (AGI). With increasing demands for higher-level semantics and cross-modal fusion, multimodal large language models (MLLMs) have emerged, integrating diverse information sources (e.g., text, vision, and audio) to enhance modeling and reasoning in complex scenarios. In AI for Science, multimodal emotion recognition and reasoning has become a rapidly growing frontier. While LLMs and MLLMs have achieved notable progress in this area, the field still lacks a systematic review that consolidates recent developments. To address this gap, this paper provides a comprehensive survey of LLMs and MLLMs for emotion recognition and reasoning, covering model architectures, datasets, and performance benchmarks. We further highlight key challenges and outline future research directions, aiming to offer researchers both an authoritative reference and practical insights for advancing this domain. To the best of our knowledge, this paper is the first attempt to comprehensively survey the intersection of MLLMs with multimodal emotion recognition and reasoning. The summary of existing methods mentioned is in our Github: \href{https://github.com/yuntaoshou/Awesome-Emotion-Reasoning}{https://github.com/yuntaoshou/Awesome-Emotion-Reasoning}.
\end{abstract}

\ccsdesc[500]{General and reference~Surveys and overviews}
\ccsdesc[300]{Human-centered computing~Natural language interfaces}

\keywords{Multi-modal emotion recognition, Deep Learning, Multimodal large language models, Artificial general intelligence}


\maketitle

\section{Introduction}

In recent years, large language models (LLMs) \cite{chang2024survey, xu2023large} have made significant progress in natural language processing (NLP). By expanding the scale of training data and the number of model parameters, LLMs have demonstrated unprecedented emergent capabilities, enabling them to perform well in many tasks, especially instruction following (IF) \cite{wang2023far}, in-context learning (ICL) \cite{wan2023gpt}, and chain-of-thought (CoT) \cite{wei2022chain}. IF enables the model to understand and perform complex tasks, ICL enables the model to flexibly handle different problems based on context without explicit training, and chain-of-thought enhances the model's decision-making process through step-by-step reasoning.

Although LLMs have performed well in many NLP tasks \cite{li2018deep} and have also demonstrated amazing zero-shot and few-shot reasoning capabilities in some complex real-world applications \cite{zhao2024panel, xie2023harnessing, gong2023iearth}, LLMs are essentially still "blind" to visual information. LLM working principle is mainly based on text data and cannot directly process multimodal data (e.g., images or videos). In contrast, large visual models (LVMs) can efficiently process and understand image content \cite{kirillov2023segment, shen2024aligning}. Through convolutional neural networks (CNNs) \cite{kim2014convolutional, shou2023comprehensive, shou2025masked} and Transformers \cite{ma2023transformer, meng2024deep} architectures, LVMs have demonstrated excellent performance in visual recognition and image generation \cite{liu2025llm4gen, zhu2025multibooth}. However, despite their powerful ability to reason about visual information, LVMs have certain limitations in natural language understanding and generation, resulting in lagging or lack of flexibility in reasoning. Given the outstanding performance and complementarity of LLMs and LVMs in their respective fields, combining the advantages of both has become a hot topic of research, giving rise to the emerging field of multimodal large language models (MLLMs) \cite{li2025chemvlm, yang2025mm}. Specifically, MLLMs are designed to receive, reason about, and output information from multiple modalities, such as text, images, audio, etc. Through cross-modal fusion, MLLMs enable models to process and understand more complex and diverse data, reason under multimodal inputs, and provide more accurate and rich outputs \cite{lin2025boosting}. The development of MLLM provides new perspectives and approaches for realizing true artificial intelligence, especially in tasks that require simultaneous understanding of language and visual information (e.g., visual question answering, image description generation, etc.), showing great potential and application value \cite{gao2025aim, park2025convis}.

Multimodal emotion recognition and reasoning, as a challenging task, requires not only the model to extract emotion information from a single modality, but also to perform deep reasoning in multimodal interactions to understand and capture complex emotion expressions and contexts \cite{zhao2025r1, yang2025omni, li2025multimodal, lin2025mind}. With the rapid advancement of MLLMs, the solutions to multimodal emotion recognition and reasoning tasks have also changed significantly \cite{cheng2024emotion, cheng2024sztu, yang2024emollm}. Through a unified generation paradigm, MLLMs are not only able to process multimodal data, but also to integrate information between multiple modalities, significantly improving the effect of emotion recognition and reasoning. It is worth noting that some MLLMs have been able to perform multimodal emotion recognition and reasoning without any additional training data, which means that they have strong zero-shot and few-shot reasoning capabilities \cite{xu2024multimodal, belikova2024deeppavlov}. Compared with traditional multimodal emotion recognition models, the latter usually rely on supervised learning and require a large amount of labeled data to fine-tune the model to adapt to different emotion recognition tasks \cite{meng2024masked, shou2024adversarial, shou2024low, ai2024gcn, shou2025spegcl, shou2024efficient}. MLLMs are able to surpass traditional multimodal emotion recognition models without the need for large-scale labeled data, which gives them a significant advantage in multimodal emotion recognition and reasoning tasks. More importantly, MLLMs can share knowledge across multiple modalities and process multiple data sources through joint training and reasoning, which provides stronger reasoning ability and higher accuracy. 

Due to the remarkable progress of LLMs and MLLMs in emotion recognition and reasoning, the interest and investment in these models in academia and industry have also shown a sharp growth trend. Therefore, this paper aims to explore the following key questions: (1) What is the current application status of LLMs and MLLMs in emotion recognition and reasoning tasks? We will review the relevant literature and analyze the specific usage and advantages of LLMs and MLLMs in emotion recognition and reasoning. (2) Have traditional multimodal emotion recognition methods been replaced by MLLMs, or can they effectively make up for the shortcomings of traditional methods? (3) What is the future development direction of MLLMs for multimodal emotion recognition and reasoning?

\begin{figure}
	\centering
	\includegraphics[width=1\linewidth]{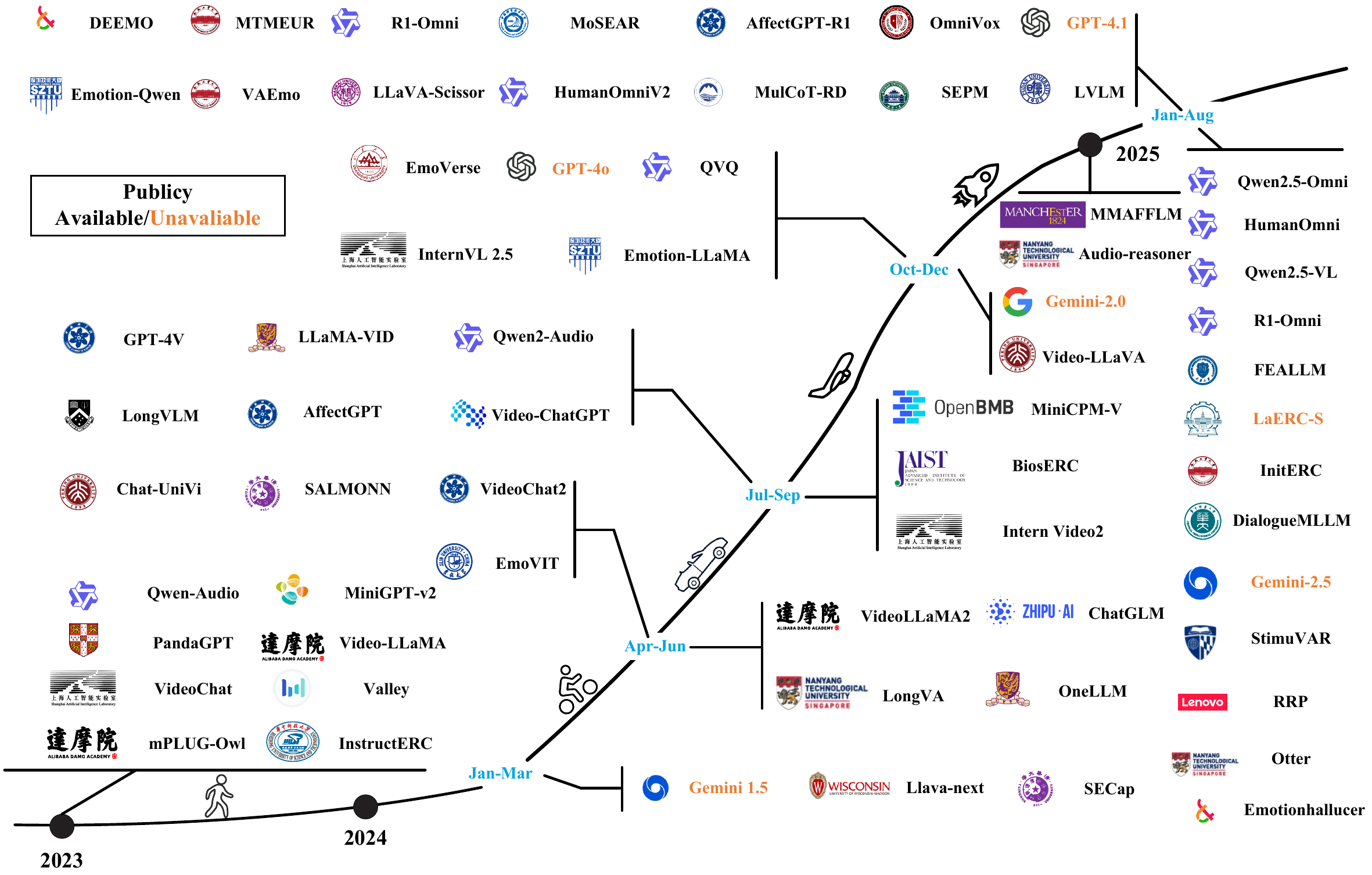}
	\caption{A chronological overview of representative MLLMs is presented, highlighting the rapid growth of this field. Additional works are continuously collected and made available on our GitHub repository, which is updated on a daily basis.}
	\label{fig:llmhis}
	\vspace{-4mm}
\end{figure}

To answer the above questions, we made the first attempt to conduct a comprehensive and detailed survey of LLMs and MLLMs for emotion recognition and reasoning as shown in Fig. \ref{fig:llmhis}. The goal of this study is to explore and summarize the current development and applications of LLMs and MLLMs in emotion recognition and reasoning tasks. In view of the rapid development of this field, our study not only aims to clarify the specific applications of these models in emotion recognition and reasoning, but also hopes to reveal the potential, limitations and future research directions of these models through systematic analysis. To this end, this paper first introduces the basic concepts and background of multimodal emotion recognition and reasoning, and reviews the preliminary related research results. Then, we focus on introducing the application paradigms of LLMs and MLLMs in emotion recognition and reasoning, and propose a unified framework to help understand how MLLMs process and reason about cross-modal data. Specifically, we classify the existing application paradigms of LLMs and MLLMs into two categories. (1) \textbf{\textit{Parameter freezing applications:}} LLMs and MLLMs pre-trained models are able to perform zero-shot learning (ZSL) \cite{wang2024llms, bao2024prompting, ji2024zero} and few-shot learning (FSL) \cite{han2024few, li2024flexkbqa, cahyawijaya2024llms} without much additional data, thanks to their strong language understanding capabilities. By freezing most of the model parameters and only adjusting a few key parameters, this approach greatly improves the efficiency of the model and reduces the need for labeled data \cite{chen2025knowledge, tai2024link, wang2025few}. (2) \textbf{\textit{Parameter tuning applications:}} LLMs and MLLMs require parameter tuning on specific tasks to further improve the accuracy and adaptability of the model \cite{zhou2024empirical, wang2024m2pt}. The Full Parameter Tuning method involves fine-tuning the parameters of the entire model to ensure that it can better handle modal data in emotion recognition tasks \cite{liu2024moe, zhang2024automated}. Efficient Parameter Tuning focuses on using optimization algorithms (e.g., learning rate adjustment, progressive training, etc.) to effectively adjust a few parameters of the model with less training data and computing resources, thereby improving performance under resource-constrained conditions \cite{li2024facial}. Finally, we explore the current research bottlenecks and challenges, and conclude by identifying potential frontiers for future research, as well as relevant challenges that inspire further exploration.

In summary, our contributions are as follows:

\begin{itemize}
	\item \textbf{New Taxonomy:} We provide a new taxonomy including (1) parameter-frozen and (2) parameter-tuning methods, which provides a unified view to understand LLMs and MLLMs for emotion recognition and reasonging. To the best of our knowledge, we present the first comprehensive review of LLMs and MLLMs for emotion recognition and reasoning.
	
	\item \textbf{Comprehensive Review:} This paper provides a comprehensive review of MLLMs for multi-modal emotion recognition. For each modeling approach, we deeply analyze the application of MLLMs in multi-modal emotion reasoning and conduct a model comparison to evaluate the advantages of different methods when dealing with cross-modal data.
	
	\item \textbf{Abundant Resources:} We have collected the latest resources on emotion recognition, covering the most advanced LLMs and MLLMs. By systematically reviewing existing model architectures, datasets, and performance evaluations, this paper provides a comprehensive guide to help researchers develop algorithms for emotion reasoning.
	
	\item  \textbf{Future Directions:} We explore the recent progress of LLMs and MLLMs as emerging research areas and deeply analyze the related challenges. This paper aims to highlight the complexity of these challenges and inspire future breakthroughs in emotion understanding and reasoning, providing inspiration and directions for further development.
\end{itemize}

\section{Preliminary}

\subsection{Definition of Multimodal Emotion Recognition and Reasoning}

\textbf{Multimodal Emotion Recognition.} Multimodal emotion recognition aims to infer human emotional states by jointly leveraging heterogeneous signals such as speech, text, facial expressions, and physiological cues \cite{shou2022conversational, shou2024adversarial, shou2025contrastive, ai2025revisiting}. Unlike unimodal recognition, multimodal emotion recognition integrates complementary and redundant information across modalities, thereby enabling more robust and reliable affective inference \cite{ai2024gcn}. Formally, given an input set of $M \in \{1, 2, \ldots m\}$ modalities as follows:
\begin{equation}
	\mathcal{X} = \{X^{(1)}, X^{(2)}, \ldots, X^{(m)}\}
\end{equation}
where each modality $X^{(m)} \in \mathbb{R}^{d_n \times T}$ represents a temporal sequence of dimension $d_n$ over $T$ time steps, the task of multimodal emotion recognition is to learn a mapping function $f: \mathcal{X} \rightarrow \mathcal{Y}$, where $\mathcal{Y}$ denotes the label space of emotions (categorical, dimensional, or mixed).

To capture inter-modal and intra-modal dependencies, multimodal emotion recognition models often introduce a latent joint representation $\mathbf{z}$ that integrates modality-specific embeddings:
\begin{equation}
	\mathbf{z} = \mathcal{F}\left(\phi_1(X^{(1)}), \phi_2(X^{(2)}), \ldots, \phi_m(X^{(m)})\right)
\end{equation}
where $\phi_m(\cdot)$ denotes the feature extractor for modality $m$, and $\mathcal{F}(\cdot)$ is the fusion function capturing cross-modal interactions. The final prediction is obtained via:
\begin{equation}
	\hat{y} = \arg\max_{y \in \mathcal{Y}} \; P(y \mid \mathbf{z}; \theta)
\end{equation}
with $\theta$ denoting the trainable parameters of the model. Therefore, multimodal emotion recognition can be rigorously defined as the problem of learning the optimal $\theta^\ast$ such that:
\begin{equation}
	\theta^\ast = \arg\min_{\theta} \; \mathbb{E}_{(\mathcal{X}, \mathcal{Y}) \sim \mathcal{D}} \; \mathcal{L}\big(\mathcal{F}(\mathcal{X}; \theta), \mathcal{Y}\big)
\end{equation}
where $\mathcal{D}$ is the multimodal dataset and $\mathcal{L}(\cdot)$ is an emotion recognition loss (e.g., cross-entropy for classification or mean squared error for dimensional regression).

\textbf{Multimodal emotion reasoning.} While multimodal emotion recognition focuses on mapping multimodal inputs to emotion labels, multimodal emotion reasoning extends the task by generating not only predictions but also interpretable explanations or causal reasoning chains \cite{zhang2025mme, song2024mosabench}. Formally, the reasoning-enhanced task can be defined as follows:
\begin{equation}
	g: \mathcal{X} \rightarrow (\mathcal{Y}, \mathcal{R})
\end{equation}
where $\mathcal{R}$ denotes the reasoning space, consisting of causal relations, evidential supports, or natural language explanations. A typical probabilistic formulation is
\begin{equation}
	P(y, r \mid \mathcal{X}) = P(y \mid \mathcal{X}) \cdot P(r \mid y, \mathcal{X})
\end{equation}
where $y \in \mathcal{Y}$ is the predicted emotion label and $r \in \mathcal{R}$ represents its associated reasoning path. This expansion highlights the importance of transparency, explainability, and causal understanding of reasoning. This shift is crucial for building trustworthy multimodal emotion systems.

\subsection{Evolution of Traditional Multimodal Emotion Recognition Methods}

The evolution of Multimodal Emotion Recognition has been driven by the need to effectively integrate heterogeneous signals and capture both their complementary and conflicting cues. Early approaches primarily relied on \textit{feature-level fusion} \cite{chudasama2022m2fnet, zhou2021information} and \textit{decision-level fusion} \cite{zhao2021multimodal, sun2024muti}, where unimodal features were either concatenated or late-averaged to yield the final prediction:
\begin{equation}
	\hat{y} = \arg\max_{y \in \mathcal{Y}} \; P\big(y \mid [\phi_1(X^{(1)});\phi_2(X^{(2)});\ldots;\phi_m(X^{(m)})]\big)
\end{equation}
where $[\cdot]$ denotes feature concatenation and $\phi_m(\cdot)$ extracts unimodal representations. Despite their simplicity, such methods suffered from overfitting and limited capacity to model inter-modal interactions. The advent of deep learning enabled hierarchical feature learning and marked a paradigm shift from shallow fusion to \textit{representation learning-based multimodal emotion recognition}. Convolutional Neural Networks (CNNs) \cite{kim2014convolutional} have demonstrated strong capability in capturing local spatial and spectral patterns, particularly in unimodal data such as facial images and speech spectrograms. However, in multimodal emotion recognition, a critical requirement is to effectively integrate heterogeneous modalities (e.g., visual, acoustic, and physiological signals) while preserving their modality-specific structures. To this end, CNNs are extended from unimodal representation learners to \textit{multimodal fusion operators}, where modality-specific feature maps are first extracted and subsequently aligned and fused in a shared latent space. Formally, given an input feature set $\{X^{(m)}\}_{m=1}^M$ from $m$-th modality, CNNs first compute unimodal feature responses:
\begin{equation}
	Y^{(m)}_{i,j} = \sigma\left(\sum_{p=1}^{k_h}\sum_{q=1}^{k_w} K^{(m)}_{p,q} \cdot X^{(m)}_{i+p, j+q} + b^{(m)}\right)
	\label{eq:cnn_unimodal}
\end{equation}

To enable cross-modal interaction, these modality-specific features are projected into a shared embedding space via learnable transformation matrices $\mathbf{W}^{(m)}$, yielding
\begin{equation}
	Z^{(m)} = \mathbf{W}^{(m)} \cdot \text{vec}\left(Y^{(m)}\right)
	\label{eq:projection}
\end{equation}
where $\text{vec}(\cdot)$ denotes vectorization. The multimodal fusion is then performed as:
\begin{equation}
	F = \phi\left( \bigoplus_{m=1}^M Z^{(m)} \right)
	\label{eq:fusion}
\end{equation}
where $\bigoplus$ denotes a modality fusion operator (e.g., concatenation, bilinear pooling, or tensor fusion), and $\phi(\cdot)$ denotes a non-linear transformation. This fusion mechanism not only preserves \textit{intra-modal representations} but also enables \textit{cross-modal interactions}, thereby capturing emotion-relevant dependencies across modalities.

Recurrent Neural Networks (RNNs), and especially Long Short-Term Memory (LSTM) models \cite{poria2017context}, are widely adopted to capture sequential dependencies in speech and conversational data. Formally, given $m$ modalities, we first compute modality-specific gates:
\begin{equation}
	\mathbf{i}^{(m)}_t, \mathbf{f}^{(m)}_t, \mathbf{o}^{(m)}_t = \sigma\left(W^{(m)} X^{(m)}_t + U^{(m)} \mathbf{h}_{t-1} + \mathbf{b}^{(m)}\right)
\end{equation}
and candidate memory content as:
\begin{equation}
	\tilde{\mathbf{c}}^{(m)}_t = \tanh\left(W^{(m)}_c X^{(m)}_t + U^{(m)}_c \mathbf{h}_{t-1}\right)
\end{equation}

Then, multimodal fusion is introduced at the cell state level via:
\begin{equation}
	\mathbf{c}_t = \sum_{m=1}^M \alpha^{(m)} \left( \mathbf{f}^{(m)}_t \odot \mathbf{c}^{(m)}_{t-1} + \mathbf{i}^{(m)}_t \odot \tilde{\mathbf{c}}^{(m)}_t \right)
	\label{eq:mm_cell}
\end{equation}
where $\alpha^{(m)}$ are learnable or attention-based modality weights ensuring adaptive contribution from each modality. Finally, the hidden state is updated as:
\begin{equation}
	\mathbf{h}_t = \phi\left( \bigoplus_{m=1}^M \mathbf{o}^{(m)}_t \odot \tanh(\mathbf{c}_t) \right)
	\label{eq:mm_hidden}
\end{equation}

These advances enabled models to learn temporal emotion dynamics rather than relying solely on static fusion. The introduction of attention mechanisms further enhanced multimodal emotion recognition by allowing selective focus on salient cross-modal cues \cite{shou2024low, meng2024masked}. Cross-modal attention is formalized as follows:
\begin{equation}
	\alpha_{t}^{(i \rightarrow j)} = \frac{\exp \big( \mathbf{q}_t^{(i)} \cdot \mathbf{k}_t^{(j)} \big)}{\sum_{j'} \exp \big( \mathbf{q}_t^{(i)} \cdot \mathbf{k}_t^{(j')} \big)}
\end{equation}
where $\mathbf{q}_t^{(i)}$ and $\mathbf{k}_t^{(j)}$ are query and key representations from modality $i$ and $j$. Building upon this principle, Transformer architectures enabled scalable modeling of long-range dependencies through multi-head self-attention \cite{lian2021ctnet, ma2023transformer}:
\begin{equation}
	\mathrm{Attention}(\mathbf{Q}, \mathbf{K}, \mathbf{V}) = \mathrm{softmax}\left( \frac{\mathbf{Q}\mathbf{K}^\top}{\sqrt{d_k}} \right)\mathbf{V}
\end{equation}
where $\mathbf{Q}$, $\mathbf{K}$, $\mathbf{V}$ denoting query, key, and value matrices. Transformers have since become a dominant backbone in multimodal emotion recognition due to their ability to align asynchronous multimodal signals at scale. To extend attention into multimodal contexts, modality-specific queries, keys, and values 
$\{\mathbf{Q}^{(m)}, \mathbf{K}^{(m)}, \mathbf{V}^{(m)}\}_{m=1}^M$ are first derived.  
Then, cross-modal attention aggregates complementary signals by computing:
\begin{equation}
	Z^{(i)}_t = \sum_{j=1}^M \alpha^{(i \rightarrow j)}_t \mathbf{V}^{(j)}_t
	\label{eq:cm_attention}
\end{equation}
This allows each modality $i$ to selectively attend to relevant information from other modalities $j$.  

Finally, the fused multimodal representation at time $t$ is obtained as
\begin{equation}
	F_t = \phi\left(\bigoplus_{i=1}^M \mathbf{Z}^{(i)}_t \right)
	\label{eq:final_fusion}
\end{equation}
This formulation enables both intra-modal temporal modeling and cross-modal interaction, thereby aligning asynchronous multimodal signals and enhancing emotion-related representation learning.

Beyond sequence-based modeling, graph neural networks (GNNs) advanced multimodal emotion recognition by representing multimodal interactions as structured graphs \cite{meng2024multi, shou2025dynamic, shou2024graph, ijcai2025p688}. Given a multimodal graph $\mathcal{G} = (\mathcal{V}, \mathcal{E})$, with nodes denoting modality-specific features and edges encoding inter-modal relations, message passing iteratively refines node embeddings. To explicitly capture cross-modal interactions, we extend the graph to a multimodal setting. Each node is associated with a modality index $m \in \{\text{audio}, \text{text}, \text{video}\}$ and initialized with modality-specific feature $\mathbf{h}_v^{(0,m)}$.  
During message passing, cross-modal attention weights are introduced to adaptively fuse signals from different modalities:
\begin{equation}
	\mathbf{h}_v^{(l+1,m)} = \sigma \left( \sum_{u \in \mathcal{N}(v)} \sum_{n=1}^M \alpha_{uv}^{(m \leftarrow n)} W^{(l,m,n)} \mathbf{h}_u^{(l,n)} \right)
	\label{eq:mm_gnn}
\end{equation}
where $\mathcal{N}(v)$ is the neighborhood of node $v$, $\alpha_{uv}^{(m \leftarrow n)}$ denotes the attention coefficient that controls the contribution of modality $n$ at node $u$ to modality $m$ at node $v$.  
The final multimodal embedding is obtained by fusing across modalities:
\begin{equation}
	Z_v = \phi \left( \bigoplus_{m=1}^M \mathbf{h}_v^{(L,m)} \right)
	\label{eq:gnn_fusion}
\end{equation}

This formulation enables GNN-based multimodal emotion recognition to (i) preserve modality-specific structural information, (ii) adaptively integrate complementary cues through cross-modal message passing, and (iii) mitigate modality imbalance and redundancy by dynamically weighting heterogeneous edges.

\subsection{LLM-based Methods}
The first stage of multimodal emotion recognition and reasoning primarily relied on large language models (LLMs) as the central reasoning engine \cite{lei2023instructerc}. In this paradigm, modality-specific encoders are employed to extract unimodal features. For example, text inputs are processed by a pre-trained text encoder to obtain embeddings. To leverage the powerful reasoning capability of LLMs, these embeddings are further projected into a unified textual space before being fed into the LLM, as illustrated in Fig.~\ref{fig:llmmllm}(a). Formally, let $\phi_m(\cdot)$ denote the encoder for modality $m \in \{\text{text}\}$, the embedding is defined as:
\begin{equation}
	z_m = \phi_m(X^{(m)})
\end{equation}

The final prediction is then generated by the LLM as:
\begin{equation}
	\hat{y} = \arg\max_{y \in \mathcal{Y}} P\big(y \mid z_m\big)
\end{equation}

Despite their simplicity, such pipelines suffer from one main limitations: the reliance on textual features alone and the inability to capture fine-grained cross-modal interactions, which are particularly crucial for affective understanding. As a result, although LLM-based methods established the feasibility of leveraging language models for emotion recognition, their scalability and reasoning fidelity remain limited.

\subsection{MLLM-based Methods}
\begin{figure}
	\centering
	\includegraphics[width=1\linewidth]{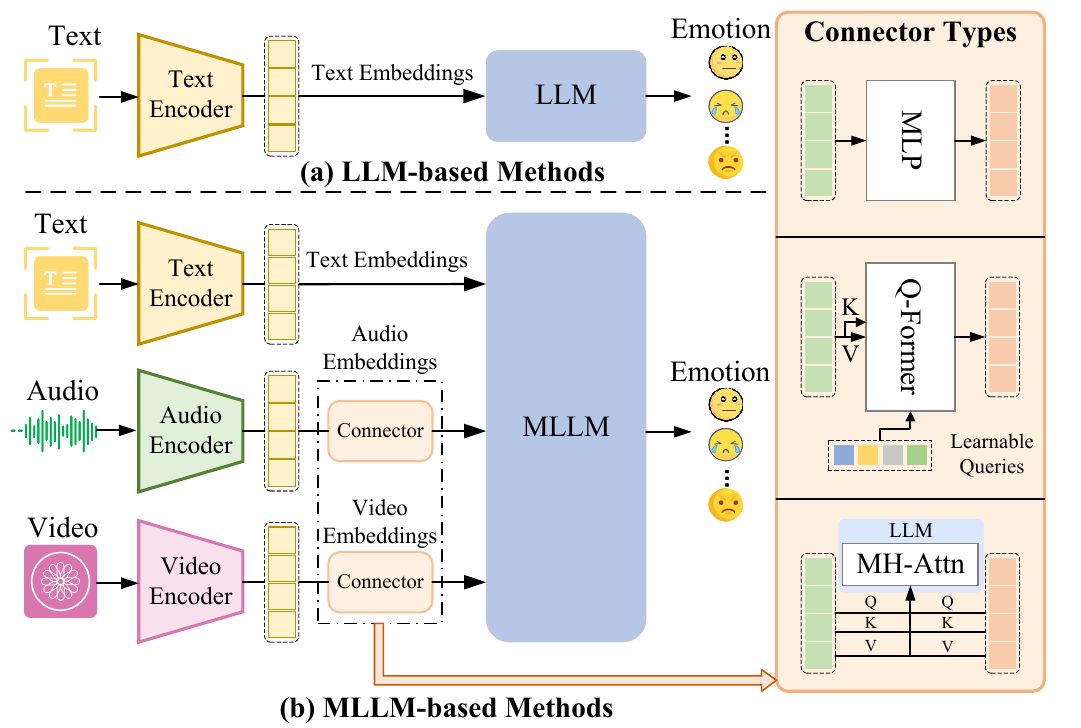}
	\caption{Comparison between LLM-based and MLLM-based methods for multimodal emotion recognition and reasoning. (a) LLM-based pipelines rely on modality-specific encoders whose outputs are projected into a textual space before being processed by the LLM, often leading to information loss and failing to enable cross-modal interaction. (b) MLLM-based architectures introduce connector modules (e.g., MLP projection, Q-Former, cross-attention) that align heterogeneous embeddings with the token space of the LLM, enabling end-to-end multimodal reasoning and richer cross-modal integration.}
	\label{fig:llmmllm}
	\vspace{-4mm}
\end{figure}

The emergence of multimodal large language models (MLLMs) marks a fundamental paradigm shift in multimodal emotion recognition and reasoning \cite{yang2025omni, DialogueMLLM}. Unlike LLM-based pipelines that retrofit unimodal embeddings into the textual space, MLLMs are inherently designed to integrate heterogeneous signals in an end-to-end manner. As shown in Fig.~\ref{fig:llmmllm}(b), embeddings from text, audio, and video encoders are aligned with the token space of the language model through connector modules, which serve as the critical bridge between modality encoders and the generative reasoning capability of LLMs. The design of these connectors has gradually evolved to support more expressive multimodal alignment \cite{wang2024llms}. The simplest approach employs a multilayer perceptron (MLP) to linearly project modality embeddings into the token dimension of the LLM, providing a lightweight yet effective alignment mechanism. To further enhance cross-modal interaction, Q-Former \cite{li2023blip} structures have been introduced, which rely on a set of learnable queries to selectively extract salient semantic features from modality embeddings before injecting them into the LLM, thereby reducing redundancy while preserving task-relevant cues. More recent works adopt cross-attention mechanisms, where the LLM directly attends to non-textual embeddings through multi-head attention layers, enabling dynamic, context-dependent fusion that better captures the complementary or conflicting nature of multimodal signals. Formally, given encoder outputs $z_m = \phi_m(X^{(m)})$, a connector $g(\cdot)$ transforms them into aligned representations:
\begin{equation}
	\tilde{z}_m = g(z_m), \quad m \in M
\end{equation}
These aligned embeddings are then jointly consumed by the MLLM for reasoning:
\begin{equation}
	\hat{y} = \arg\max_{y \in \mathcal{Y}} P\big(y \mid [\tilde{z}_t, \tilde{z}_a, \tilde{z}_v]\big)
\end{equation}

This connector-centric design allows MLLMs to preserve modality-specific information while performing end-to-end multimodal reasoning, thereby offering a more faithful and semantically grounded approach for emotion recognition and affective reasoning tasks, where subtle interdependencies across language, prosody, and visual cues are essential.

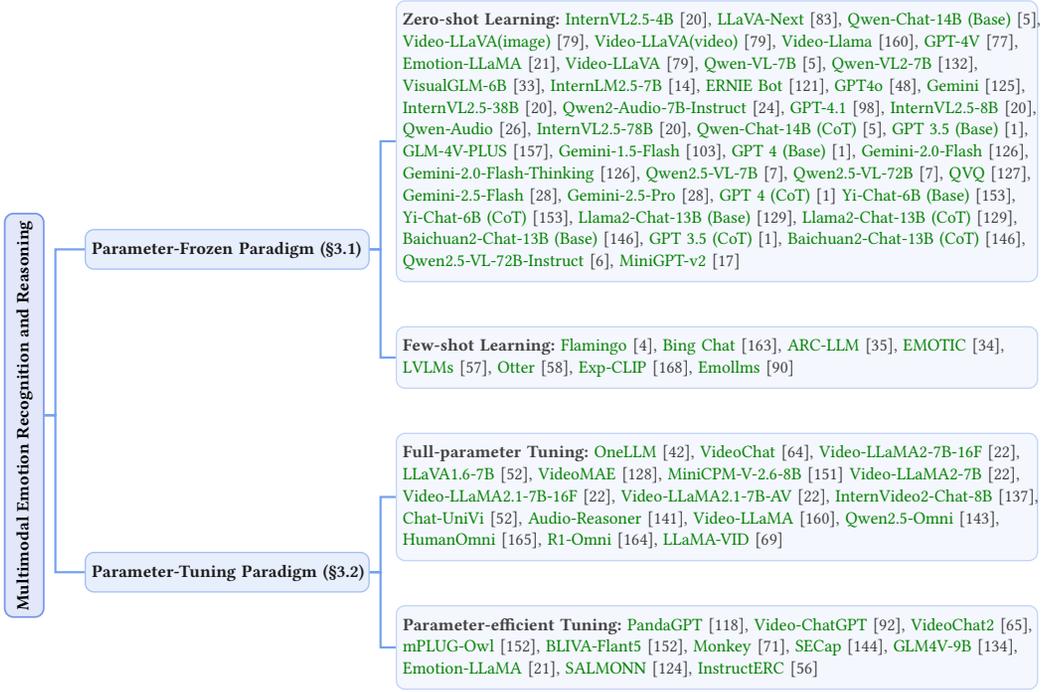
\begin{figure*}[!htbp]
	\centering
	\resizebox{1\textwidth}{!}{
		\begin{forest}
			forked edges,
			for tree={
				grow=east,
				reversed=true,
				anchor=base west,
				parent anchor=east,
				child anchor=west,
				base=left,
				font=\large\color{Black!90},      
				rectangle,
				draw=RoyalBlue!70,                 
				fill=RoyalBlue!10,                 
				rounded corners=4pt,                
				align=left,
				minimum width=2em,
				edge+={CornflowerBlue!90, line width=1pt}, 
				s sep=20pt,
				l sep=12pt,
				inner xsep=3pt,
				inner ysep=4pt,
				line width=0.6pt,
				ver/.style={
					rotate=90,
					child anchor=north,
					parent anchor=south,
					anchor=center,
					fill=RoyalBlue!15,             
					draw=RoyalBlue!70,
				},
			},
			where level=0{                          
				fill=RoyalBlue!25,
				draw=RoyalBlue!90,
				thick,
				font=\footnotesize\bfseries\color{Black},
			}{},
			where level=1{                          
				text width=12.5em,
				font=\footnotesize\bfseries\color{Black!90},
				fill=CornflowerBlue!15,
				draw=CornflowerBlue!60,
			}{},
			where level=2{                          
				text width=29em,
				font=\footnotesize\color{Black!75},
				fill=RoyalBlue!5,
				draw=CornflowerBlue!40,
			}{},
			[
			Multimodal Emotion Recognition and Reasoning, ver
			[
			Parameter-Frozen Paradigm (\S \ref{sec:Frozen})
			[
			\textbf{Zero-shot Learning:} 
			\textcolor{Green}{InternVL2.5-4B}~\cite{InternVL}{\char44}
			\textcolor{Green}{LLaVA-Next}~\cite{liu2024llava}{\char44}
			\textcolor{Green}{Qwen-Chat-14B (Base)}~\cite{bai2023qwen}{\char44}
			 \\
			\textcolor{Green}{Video-LLaVA(image)}~\cite{lin2024video}{\char44} 
			\textcolor{Green}{Video-LLaVA(video)}~\cite{lin2024video}{\char44}
			\textcolor{Green}{Video-Llama}~\cite{zhang2023video}{\char44}
			\textcolor{Green}{GPT-4V}~\cite{lian2024gpt}{\char44} \\
			\textcolor{Green}{Emotion-LLaMA}~\cite{cheng2024emotion}{\char44} 
			\textcolor{Green}{Video-LLaVA}~\cite{lin2024video}{\char44}
			\textcolor{Green}{Qwen-VL-7B}~\cite{bai2023qwen}{\char44}
			\textcolor{Green}{Qwen-VL2-7B}~\cite{wang2024qwen2}{\char44} \\
			\textcolor{Green}{VisualGLM-6B}~\cite{du2022glm}{\char44}
			\textcolor{Green}{InternLM2.5-7B}~\cite{cai2024internlm2}{\char44}
			\textcolor{Green}{ERNIE Bot}~\cite{sun2020ernie}{\char44}
			\textcolor{Green}{GPT4o}~\cite{hurst2024gpt}{\char44} 
			\textcolor{Green}{Gemini}~\cite{team2023gemini}{\char44} \\
			\textcolor{Green}{InternVL2.5-38B}~\cite{InternVL}{\char44}
			\textcolor{Green}{Qwen2-Audio-7B-Instruct}~\cite{Qwen2-Audio}{\char44}
			\textcolor{Green}{GPT-4.1}~\cite{Gpt-4.1}{\char44}
			\textcolor{Green}{InternVL2.5-8B}~\cite{InternVL}{\char44} \\
			\textcolor{Green}{Qwen-Audio}~\cite{chu2023qwen}{\char44} 
			\textcolor{Green}{InternVL2.5-78B}~\cite{InternVL}{\char44} 
			\textcolor{Green}{Qwen-Chat-14B (CoT)}~\cite{bai2023qwen}{\char44} 
			\textcolor{Green}{GPT 3.5 (Base)}~\cite{achiam2023gpt}{\char44} \\
			\textcolor{Green}{GLM-4V-PLUS}~\cite{GLM-4}{\char44}
			\textcolor{Green}{Gemini-1.5-Flash}~\cite{Gemini}{\char44} 
			\textcolor{Green}{GPT 4 (Base)}~\cite{achiam2023gpt}{\char44}
			\textcolor{Green}{Gemini-2.0-Flash}~\cite{Gemini2}{\char44} \\
			\textcolor{Green}{Gemini-2.0-Flash-Thinking}~\cite{Gemini2}{\char44}
			\textcolor{Green}{Qwen2.5-VL-7B}~\cite{bai2025qwen2}{\char44} 
			\textcolor{Green}{Qwen2.5-VL-72B}~\cite{bai2025qwen2}{\char44} 
			\textcolor{Green}{QVQ}~\cite{qvq-72b-preview}{\char44} \\
			\textcolor{Green}{Gemini-2.5-Flash}~\cite{comanici2025gemini}{\char44}
			\textcolor{Green}{Gemini-2.5-Pro}~\cite{comanici2025gemini}{\char44}
			\textcolor{Green}{GPT 4 (CoT)}~\cite{achiam2023gpt}
			\textcolor{Green}{Yi-Chat-6B (Base)}~\cite{young2024yi}{\char44} \\
			\textcolor{Green}{Yi-Chat-6B (CoT)}~\cite{young2024yi}{\char44}  
			\textcolor{Green}{Llama2-Chat-13B (Base)}~\cite{touvron2023llama}{\char44} 
			\textcolor{Green}{Llama2-Chat-13B (CoT)}~\cite{touvron2023llama}{\char44} \\
			\textcolor{Green}{Baichuan2-Chat-13B (Base)}~\cite{yang2023baichuan}{\char44}
			\textcolor{Green}{GPT 3.5 (CoT)}~\cite{achiam2023gpt}{\char44} 
			\textcolor{Green}{Baichuan2-Chat-13B (CoT)}~\cite{yang2023baichuan}{\char44}  \\
			\textcolor{Green}{Qwen2.5-VL-72B-Instruct}~\cite{Qwen-VL}{\char44} 
			 \textcolor{Green}{MiniGPT-v2}~\cite{chen2023minigpt}
			]
			[
			\textbf{Few-shot Learning:} 
			\textcolor{Green}{Flamingo}~\cite{alayrac2022flamingo}{\char44}
			\textcolor{Green}{Bing Chat}~\cite{zhang2024refashioning}{\char44}
			\textcolor{Green}{ARC-LLM}~\cite{feng2024affect}{\char44}
			\textcolor{Green}{EMOTIC}~\cite{etesam2024contextual}{\char44} \\
			\textcolor{Green}{LVLMs}~\cite{LVLM}{\char44}
			\textcolor{Green}{Otter}~\cite{Otter}{\char44}
			\textcolor{Green}{Exp-CLIP}~\cite{zhao2025enhancing}{\char44}
			\textcolor{Green}{Emollms}~\cite{liu2024emollms}
			]
			]
			[
			Parameter-Tuning Paradigm (\S \ref{sec:Tuning}) 
			[
			\textbf{Full-parameter Tuning:} 
			\textcolor{Green}{OneLLM}~\cite{Onellm}{\char44}  
			\textcolor{Green}{VideoChat}~\cite{li2023videochat}{\char44} 
			\textcolor{Green}{Video-LLaMA2-7B-16F}~\cite{VideoLLaMA_2}{\char44} \\
			\textcolor{Green}{LLaVA1.6-7B}~\cite{jin2024chat}{\char44}  
			\textcolor{Green}{VideoMAE}~\cite{tong2022videomae}{\char44}
			\textcolor{Green}{MiniCPM-V-2.6-8B}~\cite{MiniCPM-V}  
			\textcolor{Green}{Video-LLaMA2-7B}~\cite{VideoLLaMA_2}{\char44}  \\
			\textcolor{Green}{Video-LLaMA2.1-7B-16F}~\cite{VideoLLaMA_2}{\char44} 
			\textcolor{Green}{Video-LLaMA2.1-7B-AV}~\cite{VideoLLaMA_2}{\char44}
			\textcolor{Green}{InternVideo2-Chat-8B}~\cite{wang2024internvideo2}{\char44} \\
			\textcolor{Green}{Chat-UniVi}~\cite{jin2024chat}{\char44}
			\textcolor{Green}{Audio-Reasoner}~\cite{xie2025audio}{\char44}  \textcolor{Green}{Video-LLaMA}~\cite{zhang2023video}{\char44} 
			\textcolor{Green}{Qwen2.5-Omni}~\cite{xu2025qwen2}{\char44} \\
			\textcolor{Green}{HumanOmni}~\cite{zhao2025humanomni}{\char44} 
			\textcolor{Green}{R1-Omni}~\cite{zhao2025r1}{\char44}
			\textcolor{Green}{LLaMA-VID}~\cite{li2024llama}
			]
			[
			\textbf{Parameter-efficient Tuning:}
			\textcolor{Green}{PandaGPT}~\cite{su2023pandagpt}{\char44}
			\textcolor{Green}{Video-ChatGPT}~\cite{Video-ChatGPT}{\char44}
			\textcolor{Green}{VideoChat2}~\cite{li2024mvbench}{\char44}\\
			\textcolor{Green}{mPLUG-Owl}~\cite{ye2023mplug}{\char44}
			\textcolor{Green}{BLIVA-Flant5}~\cite{ye2023mplug}{\char44}
			\textcolor{Green}{Monkey}~\cite{li2024monkey}{\char44}
			\textcolor{Green}{SECap}~\cite{xu2024secap}{\char44}
			\textcolor{Green}{GLM4V-9B}~\cite{wang2024cogvlm}{\char44}\\
			\textcolor{Green}{Emotion-LLaMA}~\cite{cheng2024emotion}{\char44}
			\textcolor{Green}{SALMONN}~\cite{tangsalmonn}{\char44}
			\textcolor{Green}{InstructERC}~\cite{lei2023instructerc}
			]
			]
			]
		\end{forest}
	}
	\caption{\textbf{Taxonomy of Multimodal Emotion Recognition and Reasoning}. We systematically categorize emotion reasoning methods according to different parameter fine-tuning paradigm, demonstrating the state-of-the-art approaches across various parameter inputs.}
	\label{fig:video_taxonomy}
	\vspace{-4mm}
\end{figure*}

\section{Taxonomy}

As summarized in Fig.~\ref{fig:video_taxonomy}, we categorize multimodal emotion recognition and reasoning methods along two overarching paradigms: \emph{parameter-frozen} and \emph{parameter-tuning}. The former relies on zero-shot and few-shot prompting over frozen backbones, prioritizing rapid generalization with minimal compute, while the latter adapts model parameters either via full fine-tuning for maximal specialization or via parameter-efficient tuning (e.g., adapters, prefixes, LoRA) to balance accuracy and cost. 

\subsection{Parameter-Frozen Paradigm}

\label{sec:Frozen}

In the parameter-frozen paradigm, the backbone LLM/MLLM remains unchanged and task adaptation is realized purely through prompting and in-context specification \cite{wang2024m2pt, Otter}. This setting is particularly attractive for multimodal emotion recognition and reasoning because (i) it avoids expensive fine-tuning on large models, and (ii) it enables rapid transfer across datasets with heterogeneous label spaces. Let $X=\{x_t,x_a,x_v\}$ denote text, audio, and video inputs after modality-specific encoding or serialization into model-acceptable inputs, and let $\mathcal{I}$ denote a task instruction template that defines the goal (e.g., \emph{``Infer the speaker's emotion and briefly explain the evidence.''}). The model produces a distribution over label strings given the constructed prompt $\Pi(X,\mathcal{I})$:
\begin{equation}
	P_\Theta(y \mid X,\mathcal{I}) \;=\; \sum_{s \in \nu(y)} P_\Theta\!\big(s \,\big|\, \Pi(X,\mathcal{I})\big)
	\label{eq:verbalizer}
\end{equation}
where $\Theta$ are frozen parameters and $\nu(y)$ is a \emph{verbalizer} mapping each class $y$ to one or more label strings (e.g., \texttt{``happy''}, \texttt{``joyful''}). The final decision is $\hat y=\arg\max_{y} P_\Theta(y\mid X,\mathcal{I})$, optionally with length normalization when $\nu(y)$ contains multi-token strings.

\textbf{Zero-shot learning.} Zero-shot multimodal emotion recognition and reasoning relies solely on natural-language instructions without task-specific examples \cite{lian2024gpt, yang2025mm}. A well-engineered prompt serializes heterogeneous evidence into a structured context that highlights affective cues while minimizing spurious correlations. In LLM-based pipelines, non-text modalities are first converted into textual evidence snippets, e.g., prosodic summaries, facial action units, and salient visual events—through auxiliary encoders, and concatenated with raw transcripts using role tags (e.g., \texttt{<Text>}, \texttt{<Audio-Prosody>}, \texttt{<Visual-AU>}). In MLLM-based pipelines, raw embeddings are passed alongside textual tokens, but the model is still queried through instructions; hence Eq.~\eqref{eq:verbalizer} applies with $\Pi(\cdot)$ injecting modality tokens or connector-produced embeddings. To reduce prompt sensitivity and improve robustness, zero-shot systems commonly employ (i) instruction variants and prompt ensembling with majority voting or probability averaging, (ii) self-consistency with reasoning where latent rationales $r$ are sampled and marginalized as follows:
\begin{equation}
	P_\Theta(y \mid X,\mathcal{I}) \approx \sum_{r \in \mathcal{R}} P_\Theta(y \mid r,X,\mathcal{I})\, P_\Theta(r \mid X,\mathcal{I})
\end{equation}
and (iii) contextual calibration, which subtracts a prior estimated from a content-free prompt to alleviate label-word frequency bias. Zero-shot decoding can output both a label and an explanation, providing weak but valuable interpretability for auditing emotion decisions.

Representative systems have instantiated these principles in diverse ways. 
CLIP \cite{radford2021learning} pioneered large-scale contrastive training on image–text pairs, enabling zero-shot transfer simply by casting downstream tasks into natural-language prompts. Building on this idea, Qwen-Audio \cite{chu2023qwen} and Video-LLaVA \cite{lin2024video} extend frozen LLMs with lightweight modality encoders for audio and video streams, respectively, illustrating that non-text signals can be seamlessly injected into zero-shot pipelines. MiniGPT-v2 \cite{chen2023minigpt} and LLaVA-Next \cite{liu2024llava} further refine visual–language alignment by introducing enhanced connector modules, thereby improving the sensitivity of emotion-related cues under zero-shot prompting. Scaling up this paradigm, InternVL2.5 \cite{InternVL} and Qwen-VL \cite{bai2023qwen} leverage extensive multimodal pretraining to unify images and videos, demonstrating strong generalization across recognition and reasoning benchmarks. Beyond perception, reasoning-oriented variants such as GPT-4 (CoT) \cite{achiam2023gpt} highlight the benefit of integrating chain-of-thought prompting, which enhances interpretability by generating intermediate rationales without any parameter updates. 

\textbf{Few-shot learning.} Few-shot multimodal emotion recognition and reasoning augments the instruction with $k$ demonstration pairs $\{(X_i,y_i)\}_{i=1}^k$ injected into the prompt as in-context exemplars \cite{li2024flexkbqa, cahyawijaya2024llms}. The predictive distribution is
\begin{equation}
	P_\Theta(y \mid X^\ast,\mathcal{I},\mathcal{D}_k)
	= \sum_{s \in \nu(y)} P_\Theta\!\big(s \,\big|\, \Pi(X^\ast,\mathcal{I},\mathcal{D}_k)\big)
\end{equation}
where the \emph{selection} of $\mathcal{D}_k$ is crucial. Effective strategies balance relevance to the query and diversity across emotions and modalities, often via embedding-based retrieval with a determinantal or max–min objective to avoid redundancy \cite{tai2024link}. Demonstrations should expose the model to prototypical prosodic contours, facial micro-expressions, and discourse contexts (e.g., irony, negation) that are underrepresented in pre-training. Ordering also matters: placing task definition first, then structured exemplars (``context $\rightarrow$ rationale $\rightarrow$ label''), and finally the query tends to reduce hallucination. For classification with free-form generation, a rationale-then-verbalizer template first eliciting a short explanation $r$ and then constraining the final answer to $\nu(\mathcal{Y})$ often improves calibration and stability. When label spaces differ across datasets, dynamic verbalizers provide a principled bridge by defining synonyms per class and aggregating token probabilities as in Eq.~\eqref{eq:verbalizer}.

Representative few-shot systems have further substantiated these principles. Otter \cite{Otter} and BingChat \cite{zhang2024refashioning} illustrate how general-purpose large language models can be adapted to multimodal settings with only a handful of in-context exemplars, showing consistent improvements over zero-shot prompting. LVLMs \cite{LVLM} extend this approach by explicitly integrating visual evidence with frozen LLMs, thereby enhancing the model’s ability to capture subtle affective cues such as facial micro-expressions in few-shot contexts. Domain-specific adaptations like EMOTIC \cite{etesam2024contextual} demonstrate that curated multimodal demonstrations can guide models toward recognizing nuanced emotional states beyond basic categories. At a larger scale, Flamingo \cite{alayrac2022flamingo} exemplifies the state of the art in multimodal few-shot learning by introducing a gated cross-attention mechanism that allows visual tokens to interact directly with frozen LLM layers, enabling strong performance across a wide variety of benchmarks without full fine-tuning. Subsequent works such as ARC-LLM \cite{feng2024affect} and Exp-CLIP \cite{zhao2025enhancing} push this direction further by leveraging explicit rationales or structured demonstrations, improving interpretability while mitigating label imbalance. Collectively, these systems show that few-shot learning provides a pragmatic balance between efficiency and adaptability, making it an attractive paradigm for emotion recognition and reasoning where annotated data is scarce yet diverse affective cues must be reliably captured.

\subsection{Parameter-Tuning Paradigm}
\label{sec:Tuning}

Unlike the parameter-frozen setting, the parameter-tuning paradigm adapts the internal weights of LLMs/MLLMs to the target task, thereby improving domain specialization and calibration for multimodal emotion recognition and reasoning \cite{zhou2024empirical, liu2024moe}. Let $\phi_m$ be modality encoders for $m\!\in\!\{\text{text},\text{audio},\text{video}\}$, $g_\psi$ the connector(s), and $f_\Theta$ the LLMs/MLLMs with parameters $\Theta$. Given $\mathcal{D}=\{(X,y,r)\}$ where $X=\{x_t,x_a,x_v\}$, $y$ is a discrete label or continuous affect (valence–arousal), and $r$ is an optional rationale, a generic supervised objective is
\begin{equation}
	\small
	\mathcal{L}(\Theta,\psi)=
	\mathbb{E}_{(X,y,r)\sim\mathcal{D}}
	\Big[
	\underbrace{\mathcal{L}_{\text{task}}\big(f_\Theta(g_\psi(\{\phi_m(x_m)\}_m)),y\big)}_{\text{classification/regression}}+\lambda_r\underbrace{\mathcal{L}_{\text{gen}}\big(f_\Theta(\cdot),r\big)}_{\text{rationale}}
	+\lambda_a\underbrace{\mathcal{L}_{\text{align}}}_{\text{alignment}}
	+\lambda_t\underbrace{\mathcal{L}_{\text{temp}}}_{\text{consistency}}
	\Big]
	\label{eq:main_obj}
\end{equation}
where $\mathcal{L}_{\text{task}}$ is cross-entropy or MSE for continuous affect; $\mathcal{L}_{\text{gen}}$ is a token-level negative log-likelihood for explanation sequences; $\mathcal{L}_{\text{align}}$ often adopts a contrastive InfoNCE \cite{he2020momentum} between modality summaries $u_i$ and textual anchors $v_i$ as follows:
\begin{equation}
	\small
	\mathcal{L}_{\text{align}}
	=
	-\frac{1}{B}\!\sum_{i=1}^B
	\log
	\frac{\exp(\langle u_i,v_i\rangle/\tau)}
	{\sum_{j=1}^B \exp(\langle u_i,v_j\rangle/\tau)}
\end{equation}
and $\mathcal{L}_{\text{temp}}$ encourages smooth predictions across time for videos, e.g.,
\begin{equation}
	\small
	\mathcal{L}_{\text{temp}}
	=\frac{1}{T-1}\sum_{t=2}^{T}\big\|p_t-p_{t-1}\big\|_2^2
\end{equation}
where $p_t$ is the distribution at time $t$. Class imbalance is commonly handled via class-balanced weighting $\alpha_y \propto \frac{1-\beta}{1-\beta^{n_y}}$ or focal loss.

\begin{figure}
	\centering
	\includegraphics[width=1\linewidth]{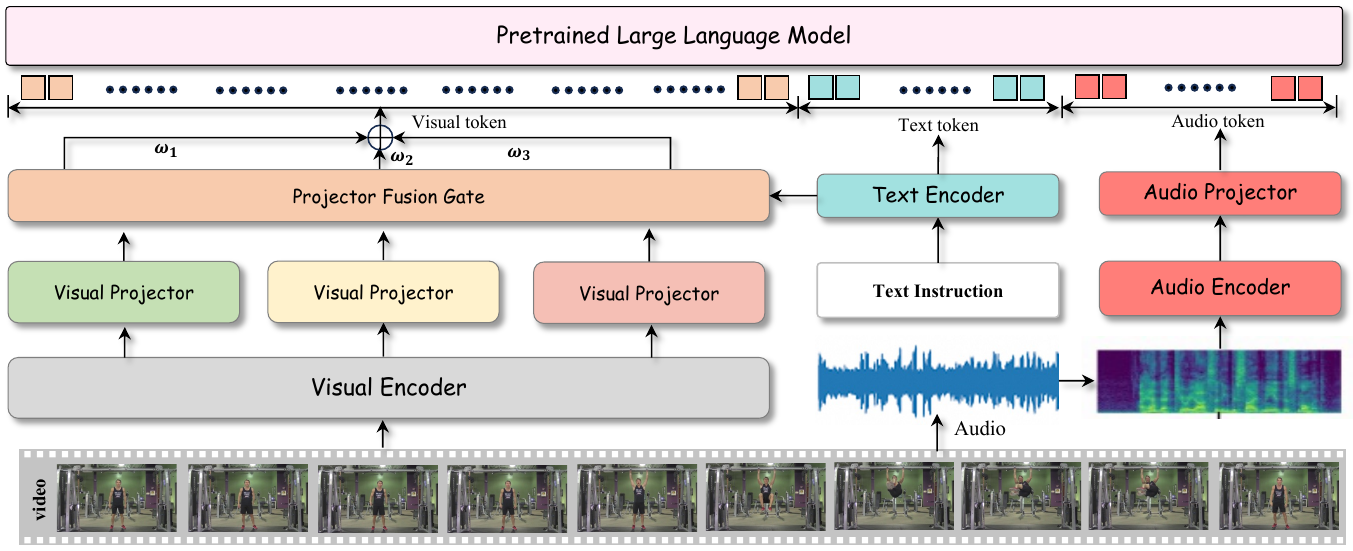}
	\caption{End-to-end full-parameter tuning architecture for multimodal emotion recognition and reasoning. Video frames are encoded by a visual encoder and mapped to the LLM token space by multiple visual projectors. A projector fusion gate with learnable coefficients $(\omega_1,\omega_2,\omega_3)$ aggregates the projected features and emits visual tokens. Audio is processed by an audio encoder and projector to produce audio tokens, while the text instruction is embedded by a text encoder. The resulting visual, audio, and text tokens are interleaved into the pretrained LLM’s token stream for joint reasoning \cite{zhao2025humanomni}.}
	\label{fig:methodvideo2}
	\vspace{-4mm}
\end{figure}

\textbf{Full-parameter tuning.} Full fine-tuning \cite{HumanOmniV2, yuan2023rrhf} updates all parameters of the backbone LLM/MLLM as shown in Fig. \ref{fig:methodvideo2}, the connectors, and optionally the modality encoders:
\begin{equation}
	\small
	(\Theta^\star,\psi^\star)=\arg\min_{\Theta,\psi}\;\mathcal{L}(\Theta,\psi)\ \ \text{s.t.}\ \ \Theta,\psi\ \text{trainable}
\end{equation}
This yields the greatest flexibility, facilitating domain-specific lexical cues, prosodic–linguistic coupling, and long-context dialogue reasoning but demands substantial memory and compute. Practical recipes include stage-wise training (first warm up $g_\psi$ and cross-attention, then unfreeze upper Transformer blocks), mixed-precision with gradient checkpointing, and curriculum schedules that progress from single-utterance supervision to multi-turn dialogue windows. Instruction tuning with multimodal emotion recognition and reasoning-style prompts further improves alignment between training signals and inference-time usage; when high-quality human preferences on explanations are available, one may add preference optimization (e.g., DPO \cite{wu2024beta}) with a pairwise objective for preferred $y^+$ over $y^-$ as follows:
\begin{equation}
	\small
	\mathcal{L}_{\text{DPO}}
	=-\mathbb{E}\big[\log\sigma\big(\beta[\log\pi_\Theta(y^+\!\mid X)-\log\pi_\Theta(y^-\!\mid X)]
	-[\log\pi_{\text{ref}}(y^+\!\mid X)-\log\pi_{\text{ref}}(y^-\!\mid X)]\big)\big]
\end{equation}

Representative full-tuning approaches have demonstrated the strengths and limitations of this paradigm. 
Early multimodal systems such as VisualBERT \cite{li2019visualbert} and UNITER \cite{chen2020uniter} established the foundation by fully fine-tuning pretrained transformers on paired image–text datasets, achieving strong performance in tasks like visual question answering and image–text retrieval. 
Building on these successes, video-oriented models such as VideoMAE \cite{tong2022videomae} and Video-LLaMA \cite{VideoLLaMA_2} extended full-parameter tuning to spatiotemporal settings, showing that large-scale video pretraining coupled with complete fine-tuning yields substantial gains in downstream multimodal reasoning tasks. OneLLM \cite{Onellm} and LLaMA-VID \cite{li2024llama} exemplify more recent efforts that scale this strategy to cover multiple modalities simultaneously, relying on end-to-end optimization to align speech, vision, and language under a unified backbone. Similarly, Chat-UniVi \cite{jin2024chat} integrates full-tuning with dialogue-style supervision, demonstrating that updating all parameters can improve long-context reasoning and multimodal dialogue alignment. 
While these models underscore the versatility and performance benefits of full-parameter tuning, they also reveal the accompanying computational burdens, motivating the exploration of more efficient alternatives such as parameter-efficient tuning.

\begin{figure}
	\centering
	\includegraphics[width=1\linewidth]{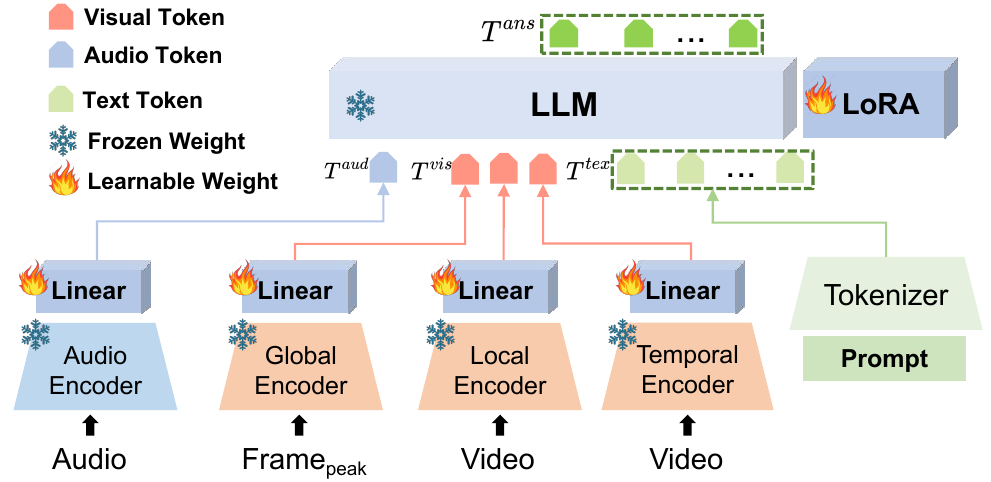}
	\caption{Parameter-efficient tuning (PET) architecture for multimodal emotion recognition and reasoning. Audio and video streams are processed by lightweight modality encoders (audio, global/local/temporal video) whose outputs are mapped to the LLM token space via shallow linear projectors to form audio and visual tokens. A tokenizer, conditioned on a textual prompt, supplies text tokens. The tokens are interleaved and fed to a pretrained LLM, where the backbone weights remain frozen while only LoRA adapters on selected layers and the linear projectors are trainable. Modality-conditioned prompt tokens $T^{\text{aud}}$, $T^{\text{vis}}$, and $T^{\text{text}}$ steer generation, and the model produces answer tokens $T^{\text{ans}}$ \cite{cheng2024emotion}.}
	\vspace{-4mm}
	\label{fig:framework}
\end{figure}

\textbf{Parameter-efficient tuning.} To lower cost while retaining most gains of full fine-tuning, parameter-efficient tuning (PET) \cite{zhou2024empirical} updates a small, carefully placed subset of parameters, keeping the base weights frozen as shown in Fig. \ref{fig:framework}. Popular instantiations include Adapter-tuning \cite{inoue2024elp}, Prefix/Prompt-tuning \cite{wang2024m2pt, wei2022chain}, and LoRA \cite{hu2022lora}/IA$^3$ \cite{liu2022few}/BitFit \cite{ben2022bitfit} families. Adapter-tuning inserts a bottleneck MLP in each block and learns only the adapter weights as follows:
\begin{equation}
	\small
	h' = h + W_{\!\uparrow}\,\sigma(W_{\!\downarrow}h),\quad
	W_{\!\downarrow}\!\in\!\mathbb{R}^{d\times r},\ 
	W_{\!\uparrow}\!\in\!\mathbb{R}^{r\times d},\ r\!\ll\! d
\end{equation}

Prefix/Prompt-tuning prepends learnable key/value vectors to each attention layer; if $K,V$ are original keys/values, the attention uses $\tilde K=[K_p;K],\ \tilde V=[V_p;V]$ with trainable prefixes $(K_p,V_p)$ while freezing backbone weights, thereby steering attention without changing $W_q,W_k,W_v$.

LoRA often with dropout on $A$ and target-specific selection of layers and learns low-rank updates for selected weight as follows:
\begin{equation}
	\small
	W' = W + \Delta W,\quad \Delta W = A B,\ \ A\!\in\!\mathbb{R}^{d\times r},\ B\!\in\!\mathbb{R}^{r\times k},\ r\!\ll\!\min(d,k)
\end{equation}

IA$^3$ applies multiplicative gating to attention/MLP projections, and BitFit updates only bias terms. For resource-constrained training, QLoRA quantizes the frozen base to 4-bit (NF4) and learns LoRA adapters in BF16/FP16, giving near full-FT performance at a fraction of memory; at inference, one can either keep the decomposition or merge $\Delta W$ into $W$ for deployment. In MLLMs, PET is frequently combined with connector tuning (train $g_\psi$ and cross-attention blocks, keep the language core frozen), or partial unfreezing of the last $L$ Transformer blocks to better adapt discourse-level reasoning while keeping compute bounded.

Representative PET-based systems have validated the practicality of this paradigm in multimodal emotion recognition and reasoning. 
For example, SECap \cite{xu2024secap} leverages selective adaptation to capture prosodic and semantic cues in speech-driven emotion captioning, significantly reducing training overhead while maintaining interpretability. 
PandaGPT \cite{su2023pandagpt} and VideoChatGPT \cite{Video-ChatGPT} extend PET strategies to video–language reasoning, integrating lightweight LoRA modules into frozen LLMs to handle long-context temporal dependencies efficiently. VideoChat2 \cite{li2024mvbench} further improves alignment by combining low-rank adapters with cross-attention, highlighting the scalability of PET for complex video–language benchmarks. 
Emotion-specific adaptations such as Emotion-LLaMA \cite{cheng2024emotion} demonstrate that parameter-efficient designs can preserve sensitivity to subtle affective states when domain-specific supervision is scarce. 
At the same time, SALMON \cite{tangsalmonn} and mPLUG-Owl \cite{ye2023mplug} show how adapter-style connectors can be tuned alongside frozen backbones to support flexible multimodal extensions. 
More recently, works like BLIVA-FlanT5 \cite{ye2023mplug}, Monkey \cite{li2024monkey}, and GLM4V-9B \cite{wang2024cogvlm} illustrate the integration of PET with instruction tuning, combining efficiency with enhanced reasoning ability. 
Together, these systems exemplify how PET techniques strike a balance between the high performance of full-parameter tuning and the scalability required for real-world deployment, underscoring their growing role in advancing MLLMs.

\begin{figure}[t]
	\centering
	\includegraphics[width=\linewidth]{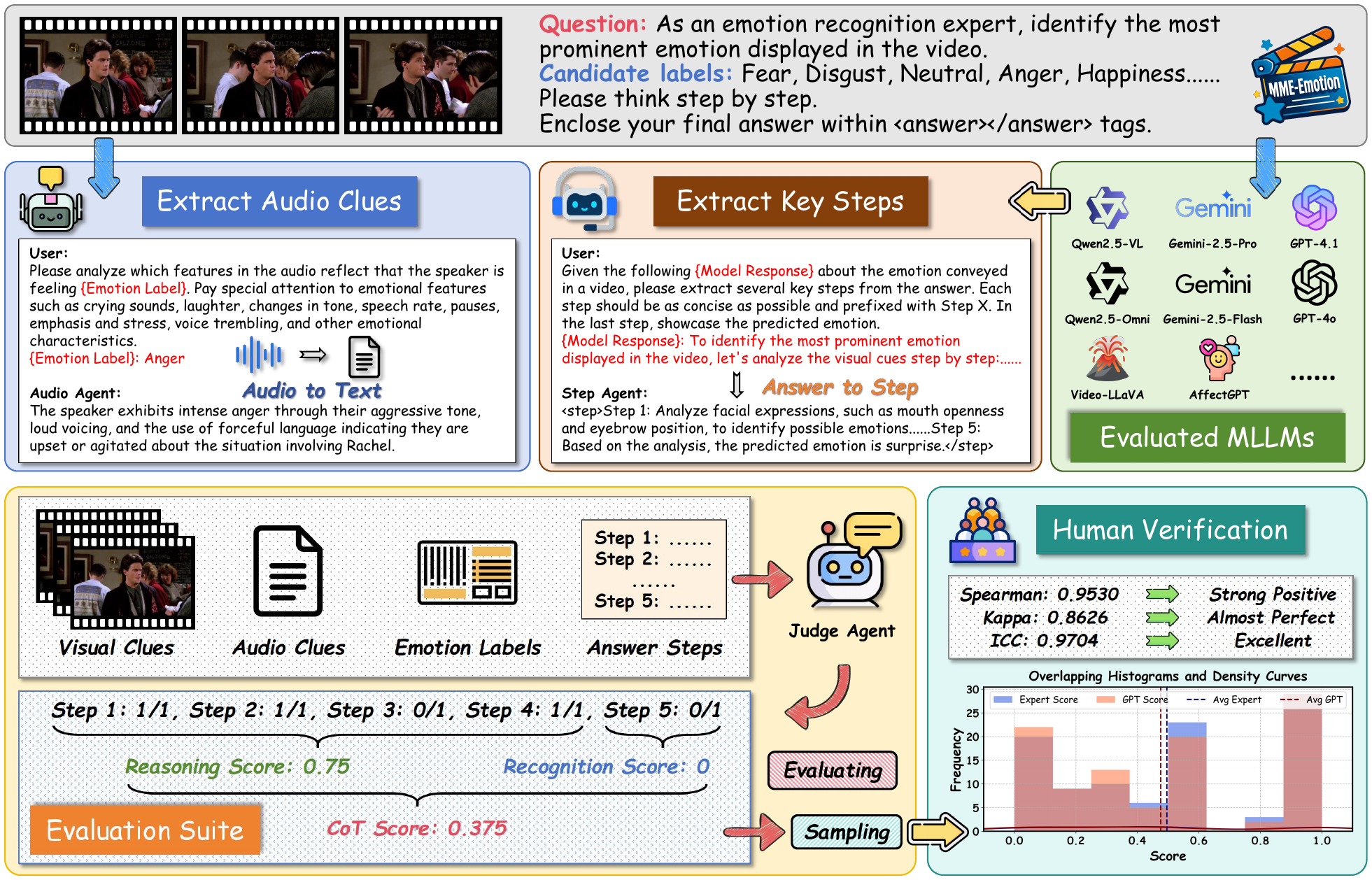}
	\caption{Pipeline for building multimodal emotion recognition and reasoning.}
	\label{fig:eval_strategy}
	\vspace{-4mm}
\end{figure}

\section{Pipeline for Constructing Emotion Recognition Datasets with MLLMs}

With the rapid advances of Multimodal Large Language Models (MLLMs), a promising paradigm has emerged to automate the construction of high-quality emotion recognition datasets. Instead of relying exclusively on labor-intensive manual annotation, MLLMs can be integrated into a multi-agent pipeline to generate emotion labels, extract rationales, and validate multimodal evidence. Fig.~6 illustrates that audio, visual, and textual cues are jointly processed to produce explainable and verifiable annotations. We summarize the overall pipeline into six key stages.

\subsection{Data Sampling and Task Formulation}
Raw sources include movies, dialogues, interviews, or spontaneous recordings. Segments are obtained by shot detection or voice activity detection (VAD) \cite{lian2024affectgpt}, followed by key-frame extraction using frame-difference, optical flow, or ORB similarity to ensure semantic diversity. Each segment is paired with an instruction template that explicitly requires step-by-step reasoning: \textit{``Identify the most prominent emotion, provide intermediate steps, and highlight modality-specific evidence.''}

\subsection{Multimodal Preprocessing}
The audio stream undergoes denoising, speaker separation, and automatic speech recognition (ASR), while extracting prosodic features such as pitch, intensity, and pauses. Visual processing involves face detection, action unit (AU) extraction, and geometric descriptors of facial regions \cite{lian2023mer, cheng2024sztu}. Temporal alignment ensures that audio, video, and transcript are synchronized into a unified multimodal sample representation.

\subsection{Generative Annotation via MLLMs}
Given the preprocessed input, multiple MLLMs (e.g., GPT, Gemini, Video-LLaMA) are prompted to output emotion labels, step-by-step rationales, and modality-specific evidence \cite{team2023gemini, achiam2023gpt, zhang2023video}. This stage converts implicit multimodal cues into explicit annotation triplets $\{ \text{label}, \text{rationales}, \text{evidence}\}$. A rationale extractor normalizes the responses into concise \textit{Step 1...N} sequences, improving consistency and interpretability.

\subsection{Judge Agent and Automated Quality Control}
To ensure annotation reliability, a \textit{Judge Agent} evaluates MLLM outputs across three dimensions: (i) recognition score $S_{\text{rec}}$ for correctness of the predicted label, (ii) reasoning score $S_{\text{reason}}$ for the logical soundness of the explanation, and (iii) chain-of-thought score $S_{\text{cot}}$ for completeness of multimodal reasoning \cite{MMEEmotion}. A unified score is computed as:
\begin{equation}
	S_{\text{final}} = \alpha \cdot S_{\text{rec}} + \beta \cdot S_{\text{reason}} + \gamma \cdot S_{\text{cot}}, \quad \alpha + \beta + \gamma = 1,
\end{equation}
with $\alpha, \beta, \gamma$ tunable across tasks. Conflicting annotations from multiple MLLMs are resolved via voting or uncertainty-aware aggregation.

\subsection{Human Verification and Consistency Checks}
Low-confidence or semantically ambiguous samples are flagged for human verification \cite{MMEEmotion, MoSEAR}. A subset of annotations is cross-evaluated by expert raters, and agreement is quantified by statistical measures such as Spearman correlation, Cohen’s $\kappa$, and the intra-class correlation coefficient (ICC). High consistency (e.g., $\kappa > 0.8$, ICC $>0.9$) indicates near-expert annotation quality, thus minimizing manual overhead.

\subsection{Structured Storage and Benchmark Release}
Finalized entries are stored in structured formats (e.g., JSON/Parquet), containing emotion labels, rationales, multimodal evidence, and quality-control metadata. The dataset supports multi-label emotions, intensity levels, and uncertainty estimates. For benchmarking, data is split into training, validation, and testing subsets, with the test set optionally hiding labels to promote explainable evaluation. Beyond accuracy and F1, evaluation protocols include rationale alignment (ROUGE/BLEU), cross-modal evidence coverage, and causal sufficiency tests \cite{lian2023explainable, lian2024affectgpt, lian2023explainable}.


\section{Popular Benchmark Datasets}

In multimodal large language model research for emotion recognition, the use of benchmark datasets is essential for algorithm validation and performance evaluation. As shown in Table \ref{tab:bench_cmp}, we summarize several commonly adopted multimodal emotion recognition datasets. Below, we provide a brief description of each dataset.

\begin{table}[t]
	\centering
	\caption{Comparison of Benchmarks related to Emotional Intelligence. Rec-A and Rea-Q are short for recognition accuracy and reasoning quality, respectively.}
	\label{tab:bench_cmp}
	\resizebox{\linewidth}{!}{%
		\begin{tabular}{l|ccccccc}
			\toprule
			\textbf{Benchmark} & \textbf{Task} & \textbf{Modality} & \textbf{Rec-A} & \textbf{Rea-Q} & \textbf{QA} & \textbf{LLM Eval} & \textbf{Human Assist} \\
			\midrule
			HUMAINE \cite{douglas2007humaine} & 1 &   Video, Audio, Text   &  \textcolor{DarkGreen}{\ding{51}}     &    \textcolor{red}{\ding{55}}  &  \textcolor{red}{\ding{55}}   &   \textcolor{red}{\ding{55}}     &  \textcolor{red}{\ding{55}}             \\
			VAM \cite{grimm2008vera}  & 1 &   Video, Audio   &   \textcolor{DarkGreen}{\ding{51}}    &   \textcolor{red}{\ding{55}}   &   \textcolor{red}{\ding{55}}  &    \textcolor{red}{\ding{55}}    &        \textcolor{red}{\ding{55}}       \\
			Youtube \cite{morency2011towards} & 1 &  Video, Audio, Text    &   \textcolor{DarkGreen}{\ding{51}}    &   \textcolor{red}{\ding{55}}   &  \textcolor{red}{\ding{55}}   &      \textcolor{red}{\ding{55}}  &    \textcolor{red}{\ding{55}}           \\
			AFEW  \cite{6200254}  & 1  &   Video, Audio   &  \textcolor{DarkGreen}{\ding{51}}     &  \textcolor{red}{\ding{55}}    &  \textcolor{red}{\ding{55}}   &      \textcolor{red}{\ding{55}}  &        \textcolor{red}{\ding{55}}       \\
			AM-FED \cite{mcduff2013affectiva}  & 1 &    Video  &   \textcolor{DarkGreen}{\ding{51}}    &   \textcolor{red}{\ding{55}}   &  \textcolor{red}{\ding{55}}   &   \textcolor{red}{\ding{55}}     &        \textcolor{red}{\ding{55}}       \\
			AFEW-VA \cite{dhall2015video}  & 1 &   Video, Audio   &  \textcolor{DarkGreen}{\ding{51}}     &  \textcolor{red}{\ding{55}}    &  \textcolor{red}{\ding{55}}   & \textcolor{red}{\ding{55}}       &     \textcolor{red}{\ding{55}}          \\
			LIRIS-ACCEDE \cite{baveye2015liris}   &  1 &   Video, Text   &  \textcolor{DarkGreen}{\ding{51}}     &   \textcolor{red}{\ding{55}}   &  \textcolor{red}{\ding{55}}   &   \textcolor{red}{\ding{55}}     &        \textcolor{red}{\ding{55}}       \\
			EMILYA \cite{fourati2014emilya} &  1  &   Video, Text   &   \textcolor{DarkGreen}{\ding{51}}    & \textcolor{red}{\ding{55}}     &   \textcolor{red}{\ding{55}}  &   \textcolor{red}{\ding{55}}     &        \textcolor{red}{\ding{55}}       \\
			SEWA \cite{kossaifi2019sewa}  &  1  &   Video, Audio   &   \textcolor{DarkGreen}{\ding{51}}    & \textcolor{red}{\ding{55}}     &  \textcolor{red}{\ding{55}}   &   \textcolor{red}{\ding{55}}     &     \textcolor{red}{\ding{55}}         \\
			CMU-MOSEI \cite{zadeh2018multimodal} & 2 &    Video, Audio, Text  &    \textcolor{DarkGreen}{\ding{51}}   &   \textcolor{red}{\ding{55}}   &  \textcolor{red}{\ding{55}}   &    \textcolor{red}{\ding{55}}    &    \textcolor{red}{\ding{55}}           \\
			iMiGUE \cite{liu2021imigue} & 1 &    Video, Text  &  \textcolor{DarkGreen}{\ding{51}}     &  \textcolor{red}{\ding{55}}    &  \textcolor{red}{\ding{55}}   &   \textcolor{red}{\ding{55}}     &     \textcolor{red}{\ding{55}}          \\
			DFEW \cite{jiang2020dfew}  &   1   &   Video, Audio, Text   &  \textcolor{DarkGreen}{\ding{51}}     &    \textcolor{red}{\ding{55}}  & \textcolor{red}{\ding{55}}    &  \textcolor{red}{\ding{55}}       &       \textcolor{red}{\ding{55}}       \\
			MER2023 \cite{lian2023mer}  &   1   &   Video, Audio, Text   &  \textcolor{DarkGreen}{\ding{51}}     &    \textcolor{red}{\ding{55}}  &  \textcolor{DarkGreen}{\ding{51}}   &  \textcolor{DarkGreen}{\ding{51}}      &      \textcolor{DarkGreen}{\ding{51}}         \\
			EMER \cite{lian2023explainable}  &   1   &   Video, Audio, Text   &  \textcolor{DarkGreen}{\ding{51}}     &    \textcolor{red}{\ding{55}}  &   \textcolor{DarkGreen}{\ding{51}}  &     \textcolor{DarkGreen}{\ding{51}}   &    \textcolor{DarkGreen}{\ding{51}}           \\
			MERR  \cite{cheng2024sztu}  &  1    &   Video, Audio, Text   &  \textcolor{DarkGreen}{\ding{51}}     &    \textcolor{red}{\ding{55}}  &   \textcolor{DarkGreen}{\ding{51}}  &    \textcolor{DarkGreen}{\ding{51}}    &     \textcolor{DarkGreen}{\ding{51}}          \\								
			DEEMO \cite{li2025deemo} &   3   &   Video, Audio, Text  &  \textcolor{DarkGreen}{\ding{51}}      &  \textcolor{red}{\ding{55}}    &  \textcolor{DarkGreen}{\ding{51}}   &    \textcolor{DarkGreen}{\ding{51}}    &      \textcolor{DarkGreen}{\ding{51}}         \\
			EmotionBench \cite{sabour2024emobench} & 3 & Text & \textcolor{DarkGreen}{\ding{51}} & \textcolor{red}{\ding{55}} & \textcolor{DarkGreen}{\ding{51}} & \textcolor{red}{\ding{55}} & \textcolor{DarkGreen}{\ding{51}} \\
			MOSABench \cite{song2024mosabench} & 1 & Image & \textcolor{DarkGreen}{\ding{51}} & \textcolor{red}{\ding{55}} & \textcolor{DarkGreen}{\ding{51}} & \textcolor{DarkGreen}{\ding{51}} & \textcolor{red}{\ding{55}} \\
			MM-InstructEval \cite{yang2025mm} & 6 & Image & \textcolor{DarkGreen}{\ding{51}} & \textcolor{red}{\ding{55}} & \textcolor{DarkGreen}{\ding{51}} & \textcolor{red}{\ding{55}} & \textcolor{red}{\ding{55}} \\
			EIBench \cite{lin2025we} & 1 &  Image & \textcolor{DarkGreen}{\ding{51}} & \textcolor{red}{\ding{55}} & \textcolor{DarkGreen}{\ding{51}} & \textcolor{DarkGreen}{\ding{51}} & \textcolor{DarkGreen}{\ding{51}} \\
			EmoryNLP \cite{zahiri2018emotion} & 1 &  Text & \textcolor{DarkGreen}{\ding{51}} & \textcolor{red}{\ding{55}} & \textcolor{red}{\ding{55}} & \textcolor{red}{\ding{55}} & \textcolor{red}{\ding{55}} \\
			IEMOCAP \cite{busso2008iemocap} & 1 &  Video, Audio, Text & \textcolor{DarkGreen}{\ding{51}} & \textcolor{red}{\ding{55}} & \textcolor{red}{\ding{55}} & \textcolor{red}{\ding{55}} & \textcolor{red}{\ding{55}} \\
			MELD \cite{poria2019meld} & 1 &  Video, Audio, Text & \textcolor{DarkGreen}{\ding{51}} & \textcolor{red}{\ding{55}} & \textcolor{red}{\ding{55}} & \textcolor{red}{\ding{55}} & \textcolor{red}{\ding{55}} \\
			MC-EIU \cite{lianov}& 2 &  Video, Audio, Text & \textcolor{DarkGreen}{\ding{51}} & \textcolor{red}{\ding{55}} & \textcolor{red}{\ding{55}} & \textcolor{red}{\ding{55}} & \textcolor{red}{\ding{55}} \\
			OV-MER \cite{lianov} & 1 &  Video, Audio, Text & \textcolor{DarkGreen}{\ding{51}} & \textcolor{red}{\ding{55}} & \textcolor{DarkGreen}{\ding{51}} & \textcolor{red}{\ding{55}} & \textcolor{DarkGreen}{\ding{51}} \\
			
			CA-MER \cite{MoSEAR} & 3 &  Video, Audio, Text & \textcolor{DarkGreen}{\ding{51}} & \textcolor{red}{\ding{55}} & \textcolor{DarkGreen}{\ding{51}} & \textcolor{DarkGreen}{\ding{51}} & \textcolor{DarkGreen}{\ding{51}} \\
			
			EmoBench-M \cite{hu2025emobench} & 3 &  Video, Audio, Text & \textcolor{DarkGreen}{\ding{51}} & \textcolor{red}{\ding{55}} & \textcolor{DarkGreen}{\ding{51}} & \textcolor{DarkGreen}{\ding{51}} & \textcolor{DarkGreen}{\ding{51}} \\
			EMER-Coarse \cite{lian2024affectgpt} & 3 &  Video, Audio, Text & \textcolor{DarkGreen}{\ding{51}} & \textcolor{red}{\ding{55}} & \textcolor{DarkGreen}{\ding{51}} & \textcolor{DarkGreen}{\ding{51}} & \textcolor{DarkGreen}{\ding{51}} \\
			
			\rowcolor{black!10} MME-Emotion \cite{zhang2025mme} & 8 & Video, Audio, Text & \textcolor{DarkGreen}{\ding{51}} & \textcolor{DarkGreen}{\ding{51}} & \textcolor{DarkGreen}{\ding{51}} & \textcolor{DarkGreen}{\ding{51}} & \textcolor{DarkGreen}{\ding{51}} \\
			\bottomrule
		\end{tabular}%
	\vspace{-4mm}
	}
\end{table}

\textbf{HUMAINE} \cite{douglas2007humaine}: The HUMAINE database is one of the earliest large-scale resources dedicated to affective computing and emotion recognition. The dataset contains both naturalistic and experimentally induced emotional expressions, spanning multiple modalities such as audio, video, and physiological signals. Importantly, it provides detailed manual annotations of emotional states based on categorical labels (e.g., joy, anger) as well as dimensional descriptors (e.g., arousal, valence).

\textbf{VAM} \cite{grimm2008vera}: The Vera am Mittag (VAM) corpus consists of spontaneous emotional expressions extracted from the German television talk show Vera am Mittag, providing a rich collection of natural audio-visual interactions. Each segment is annotated using dimensional emotion descriptors—valence, arousal, and dominance by multiple human raters, ensuring both reliability and fine-grained analysis of affective states.

\textbf{Youtube} \cite{morency2011towards}: The YouTube dataset comprises video clips collected from YouTube, capturing spontaneous expressions of opinion across diverse topics and speakers. Each clip contains audio, visual, and textual modalities, with sentiment annotations assigned at the utterance level.

\textbf{AFEW} \cite{6200254}: The Acted Facial Expressions in the Wild (AFEW) dataset is constructed from a diverse collection of movie excerpts, containing audio-visual samples that encompass a wide range of speakers, contexts, and environmental conditions such as varying lighting, and occlusions. Each sample is annotated with one of the basic emotion categories (e.g., anger, happiness, sadness). 

\textbf{AM-FED} \cite{mcduff2013affectiva}: Affectiva-MIT Facial Expression Dataset (AM-FED) contains more than 1,500 video recordings comprising over 242 hours of spontaneous facial behavior. Each video is annotated with affective labels, covering both basic emotions and continuous affective dimensions.

\textbf{AFEW-VA} \cite{dhall2015video}: The Acted Facial Expressions in the Wild–Valence and Arousal (AFEW-VA) dataset extends the original AFEW corpus by introducing continuous annotations of emotional dimensions. Specifically, it provides frame-level labels for valence and arousal in video clips extracted from movies, thereby enabling fine-grained analysis of affective dynamics.

\textbf{LIRIS-ACCEDE} \cite{baveye2015liris}: The LIRIS-ACCEDE corpus consists of more than 9,800 short movie excerpts carefully selected from Creative Commons films, covering a broad range of genres, visual styles, and emotional contexts. Each clip is annotated with continuous valence and arousal ratings.

\textbf{EMILYA} \cite{fourati2014emilya}: The EMILYA corpus contains a wide range of motion capture recordings in which actors perform daily actions under different induced emotional states such as happiness, anger, sadness, and fear. Each sequence is annotated with categorical emotion labels as well as dimensional descriptors of valence and arousal.

\textbf{SEWA} \cite{kossaifi2019sewa}: The SEWA database contains audio-visual recordings of more than 400 participants engaged in naturalistic dyadic interactions across six different cultural backgrounds. Each recording is richly annotated at multiple levels, including continuous valence, arousal, and liking dimensions. 

\textbf{CMU-MOSEI} \cite{zadeh2018multimodal}: The CMU Multimodal Opinion Sentiment and Emotion Intensity (CMU-MOSEI) dataset contains more than 23,000 annotated video segments collected from over 1,000 speakers across a wide range of topics on YouTube. Annotations cover both sentiment (from highly negative to highly positive) and six basic emotion categories, with continuous intensity scores.

\textbf{iMiGUE} \cite{liu2021imigue}: The iMiGUE comprises more than 3,000 video samples collected from over 60 participants, with deliberate anonymization to remove identity-related cues such as facial appearance. Each sequence is annotated with both categorical emotion labels and dimensional descriptors.

\begin{figure}
	\centering
	\includegraphics[width=1\linewidth]{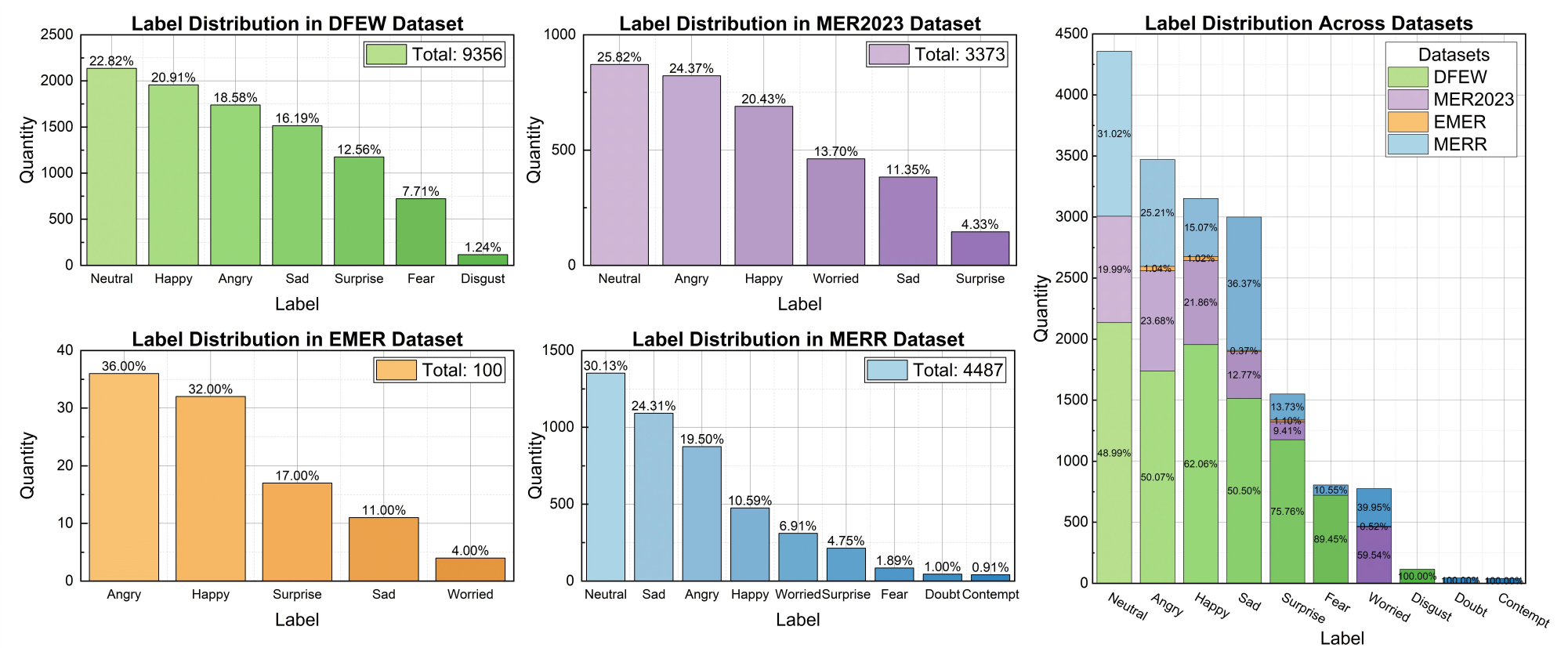}
	\caption{Comparative analysis of label distributions across the DFEW, MER2023, EMER, and
		MERRdatasets. The left side illustrates the individual label distributions for each dataset, while the right side presents a side-by-side comparison of the label distributions, highlighting the similarities and differences in emotional category representation among the datasets. \cite{cheng2024emotion}}
	\label{fig:datasetcomprision}
	\vspace{-4mm}
\end{figure}

\textbf{DFEW} \cite{jiang2020dfew}: The DFEW dataset comprises 16,372 video clips extracted from movies, capturing a wide spectrum of real-world complexities as shown in Fig. \ref{fig:datasetcomprision}. Each video clip is annotated through a rigorous crowdsourcing pipeline involving 12 expert annotators, with every clip independently labeled ten times to ensure annotation reliability. The dataset provides both single-label annotations covering the seven basic emotion categories and 7-dimensional expression distribution vectors.

\textbf{MER2023} \cite{lian2023mer}: The MER 2023 dataset was specifically designed to evaluate system robustness through three complementary tracks as shown in Fig. \ref{fig:datasetcomprision}. The MER-MULTI track requires participants to recognize both discrete categorical emotions and continuous dimensional attributes. The MER-NOISE track introduces artificial noise into test videos to systematically assess modality robustness,. The MER-SEMI track provides large volumes of unlabeled samples.

\textbf{EMER} \cite{lian2023explainable}: The Explainable Multimodal Emotion Recognition (EMER) seeks to address the inherent subjectivity and ambiguity of affective labeling as shown in Fig. \ref{fig:datasetcomprision}. Each instance in the dataset not only contains multimodal signals (text, audio, and video), but is also paired with explanatory rationales that justify the emotion annotations.

\textbf{MERR} \cite{cheng2024sztu}: The Multimodal Emotion Recognition in Real-world (MERR) dataset integrates audio, visual, and textual modalities with rich affective annotations as shown in Fig. \ref{fig:datasetcomprision}. Each instance is labeled with both categorical emotions and continuous sentiment dimensions. 

\textbf{DEEMO} \cite{li2025deemo}: The DEEMO corpus contains audio-visual-textual recordings paired with annotations covering categorical emotions, dimensional descriptors, and reasoning rationales, enabling not only recognition but also interpretability in emotion analysis.

\begin{figure}[ht]
	\centering
	\includegraphics[width=0.8\linewidth]{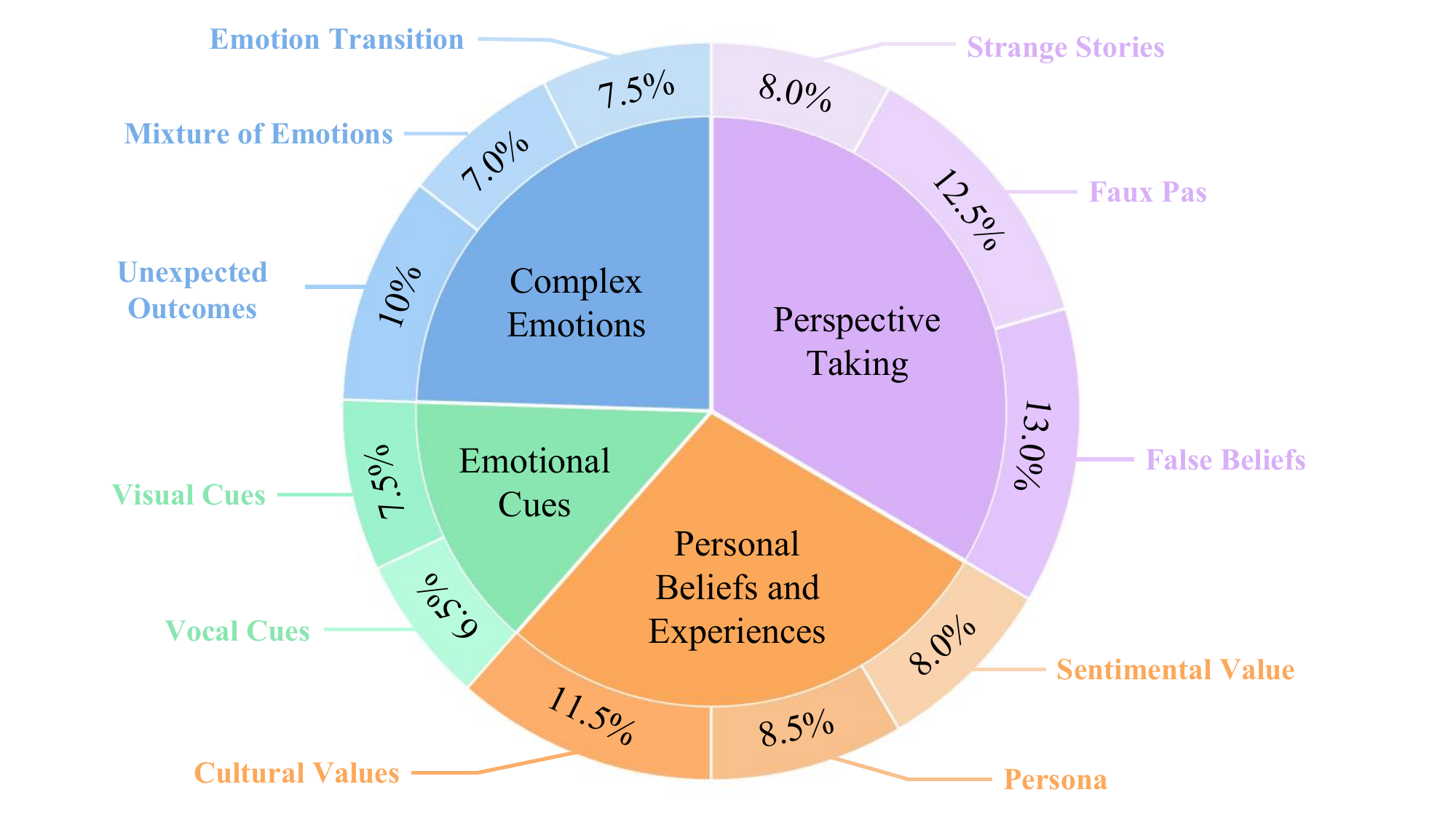}
	\caption{Category Distribution in EmotionBench. The main categories are depicted within the chart, and the secondary categories are annotated outside the chart.}
	\label{fig:topic_pic}
	\vspace{-4mm}
\end{figure}

\textbf{EmotionBench} \cite{sabour2024emobench}: The EmoBench dataset is a recently introduced benchmark specifically designed to assess the emotional intelligence of large language models (LLMs) as shown in Fig. \ref{fig:topic_pic}. EmoBench provides a comprehensive suite of emotion-related evaluations, including emotion recognition, cause identification, empathy assessment, and reasoning about affective situations. The dataset is linguistically diverse, containing carefully curated text-based scenarios that span a wide spectrum of emotional states and social contexts.

\textbf{MOSABench} \cite{song2024mosabench}: The MOSABench corpus is designed to evaluate the fine-grained sentiment understanding capabilities of multimodal large language models (MLLMs). MOSABench emphasizes complex real-world scenes containing multiple objects and entities, each associated with distinct sentiment cues. The dataset pairs images with structured annotations covering object-level sentiment labels, relational context, and overall scene-level affect.

\textbf{MM-InstructEval} \cite{yang2025mm}: The MM-InstructEval benchmark is designed to assess the multimodal reasoning capabilities of large language models under zero-shot settings. It encompasses a broad spectrum of tasks that require integrating and reasoning over textual and visual modalities, including visual question answering, commonsense reasoning with images, multimodal entailment, and instruction following grounded in visual context.

\textbf{EIBench} \cite{lin2025we}: The EIBench benchmark is specifically designed to evaluate the emotional intelligence of multimodal large language models (MLLMs). EIBench emphasizes higher-level emotional reasoning, including understanding the causes of emotions, interpreting multimodal affective cues, and generating contextually appropriate empathetic responses. 

\begin{figure}
	\centering
	\includegraphics[width=1\linewidth]{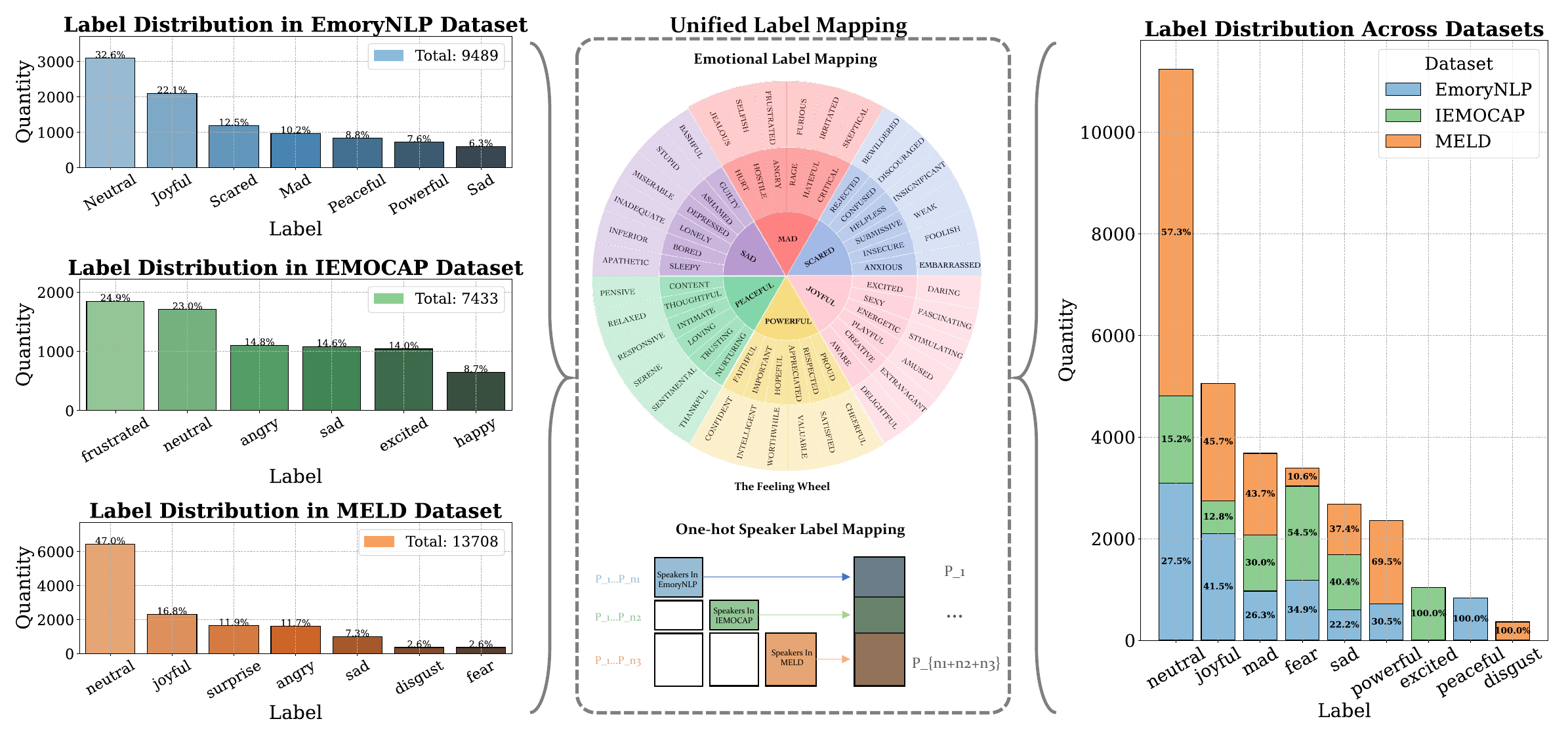}
	\caption{Unified Label Mapping Across three Open-source Benchmarks. \cite{lei2023instructerc}}
	\label{fig:unifiedlabelmappingv3}
	\vspace{-4mm}
\end{figure}

\begin{figure}[t]
	\centering
	\includegraphics[width=\linewidth]{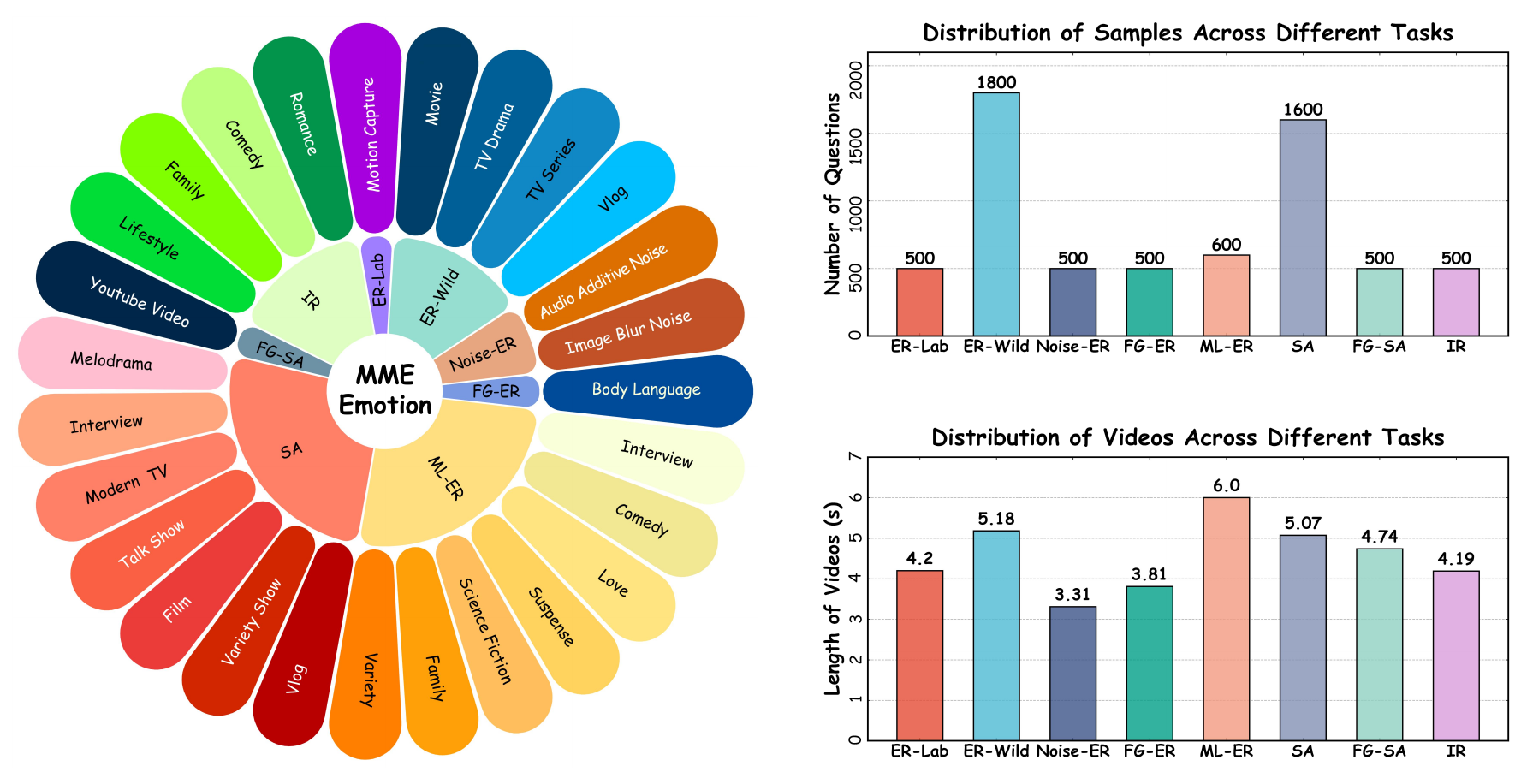}
	\caption{Overview of MME-Emotion Statistics. \textit{Left}: Task Types. MME-Emotion encompasses eight emotional tasks across 27 distinct scenario types, enabling fine-grained analysis of diverse video contexts. \textit{Right}: Data Distributions. MME-Emotion features balanced distributions of question volume and video duration, facilitating comprehensive evaluation of temporal understanding.}
	\vspace{-4mm}
	\label{fig:MME}
\end{figure}

\textbf{EmoryNLP} \cite{zahiri2018emotion}: The EmoryNLP dataset is specifically constructed to capture fine-grained affective dynamics across multi-turn dialogues as shown in Fig. \ref{fig:datasetcomprision}. Extracted from episodes of the television series Friends, it contains over 12,000 utterances annotated with seven emotion categories, including neutral, joyful, peaceful, powerful, scared, mad, and sad.

\textbf{IEMOCAP} \cite{busso2008iemocap}: The Interactive Emotional Dyadic Motion Capture (IEMOCAP) dataset contains approximately 12 hours of audiovisual recordings from ten actors engaged in both scripted and improvised dyadic interactions as shown in Fig. \ref{fig:datasetcomprision}. Each utterance is annotated with both categorical emotion labels and dimensional attributes such as valence, arousal, and dominance.

\textbf{MELD} \cite{poria2019meld}: Derived from the EmotionLines corpus, the Multimodal EmotionLines Dataset (MELD) contains more than 1,400 dialogues and approximately 13,000 utterances sourced from the television series Friends, offering naturalistic, multi-party conversational settings as shown in Fig. \ref{fig:datasetcomprision}. Each utterance is annotated with categorical emotion labels and sentiment polarity.

\textbf{MC-EIU} \cite{liu2024emotion}: The MC-EIU dataset is specifically designed to support joint modeling of emotions and intents in conversational settings. The dataset features over seven emotion categories and nine intent categories, spanning three modalities and is annotated in both English and Mandarin.

\textbf{OV-MER} \cite{lianov}: The OV-MER dataset introduces the open-vocabulary paradigm into multimodal emotion recognition (MER). Unlike traditional datasets constrained by fixed emotion taxonomies, OV-MER is designed to support flexible prediction of emotions beyond predefined categories, thereby better reflecting the complexity, subtlety, and multi-appraisal nature of human affective experiences.

\textbf{CA-MER} \cite{MoSEAR}: The CA-MER dataset is designed to address the overlooked challenge of emotion conflicts in multimodal emotion recognition. Unlike conventional datasets that assume consistency across modalities, CA-MER explicitly models scenarios where emotional cues from audio and visual streams are misaligned.

\textbf{EmoBench-M} \cite{hu2025emobench}: The EmoBench-M benchmark is designed to assess the emotional intelligence (EI) of multimodal large language models (MLLMs) in realistic interaction scenarios. Grounded in established psychological theories of EI, the dataset covers 13 evaluation scenarios spanning three critical dimensions: foundational emotion recognition, conversational emotion understanding, and socially complex emotion analysis.

\textbf{EMER-Coarse} \cite{lian2024affectgpt}: The EMER-Coarse dataset was developed as a complement to the smaller, manually verified EMER-Fine dataset, which is limited by high annotation costs. To expand coverage, EMER-Coarse adopts a simplified annotation pipeline, reduces reliance on manual verification, and leverages open-source models for large-scale automatic labeling.

\textbf{MME-Emotion} \cite{zhang2025mme}: The MME-Emotion benchmark comprises over 6,000 curated video clips paired with task-specific question–answer sets, systematically covering eight emotion-related tasks across diverse scenarios as shown in Fig. \ref{fig:MME}. MME-Emotion introduces a holistic evaluation framework that jointly assesses both emotional understanding and causal reasoning. 

\begin{table}[ht]
	\centering
	\caption{Comparison of multimodal emotion recognition results on DFEW. The upper part shows zero-shot performance, while the lower part shows results after fine-tuning.}
	\scalebox{0.95}{
		\begin{tabular}{lcccccccccc}
			\toprule
			{Method}  & {Hap} & {Sad} & {Neu} & {Ang} & {Sur} & {Dis} & {Fea} & {UAR} & {WAR}\\
			\midrule
			\textbf{\textit{Zero-Shot}} &  &   &   &   &   &   &   &   & \\
			Qwen-Audio ~\cite{chu2023qwen}          & {25.97}  & {12.93}  & {67.04}  & {29.20}  & {6.12}  & {0.00}  & {35.36}  & {25.23}  & {31.74}\\
			LLaVA-NEXT ~\cite{liu2024llava}          & {57.46}  & \textbf{79.42}  & {38.95}  & {0.00}   & {0.00}  & {0.00}  & {0.00}   & {25.12}  & {33.75}\\
			MiniGPT-v2 ~\cite{chen2023minigpt}          & \textbf{84.25}  & {47.23}  & {22.28}  & {20.69}  & {2.04}  & {0.00}  & {0.55}   & {25.29}  & {34.47}\\
			Video-LLaVA(image) ~\cite{lin2024video}  & {37.09}  & {27.18}  & {26.97}  & {58.85}  & {12.97} & {0.00}  & {3.31}   & {20.78}  & {31.10}\\
			Video-LLaVA(video) ~\cite{lin2024video}  & {51.94}  & {39.84}  & {29.78}  & {58.85}  & {0.00}  & {0.00}  & {2.76}   & {26.17}  & {35.24}\\
			Video-Llama ~\cite{zhang2023video}         & {20.25}  & {67.55}  & \textbf{80.15}  & {5.29}   & {4.76}  & {0.00}  & {9.39}   & {26.77}  & {35.75}\\
			GPT-4V ~\cite{lian2024gpt}              & {62.35}  & {70.45}  & {56.18}  & {50.69}  &{32.19} & {10.34} & \textbf{51.11}  & \textbf{47.69}      & {54.85}\\
			\rowcolor{gray!20}
			Emotion-LLaMA \cite{cheng2024emotion} & {71.98}  & {76.25}  & {61.99}  & \textbf{71.95}  & \textbf{33.67} & {0.00}  & {3.31}  & {45.59}  & \textbf{59.37}\\
			\midrule
			\textbf{\textit{Fine-tuning}} &  &   &   &   &   &   &   &   & \\
			EC-STFL ~\cite{jiang2020dfew}             & {79.18}  & {49.05}  & {57.85}  & {60.98}  & {46.15}  & {2.76}  & {21.51}  & {45.35}  & {56.51}\\
			Former-DFER ~\cite{zhao2021former}         & {84.05}  & {62.57}  & {67.52}  & {70.03}  & {56.43}  & {3.45}  & {31.78}  & {53.69}  & {65.70}\\
			IAL ~\cite{li2023intensity}    & {87.95}  & {67.21}  & {70.10}  & {76.06}  & {62.22}  & {0.00}  & {26.44}  & {55.71}  & {69.24}\\
			MAE-DFER ~\cite{sun2023mae}            & {92.92}  & {77.46}  & {74.56}  & {76.94}  & {60.99}  & \textbf{18.62} & {42.35}  & {63.41}  & {74.43}\\
			VideoMAE ~\cite{tong2022videomae}     & {93.09}  & {78.78}  & {71.75}  & {78.74}  & {63.44}  & {17.93} & {41.46}  & {63.60}  & {74.60}\\
			S2D ~\cite{chen2024static}            & \textbf{93.62}  & \textbf{80.25}  & \textbf{77.14}  & {81.09}  & {64.53}  & {1.38} & {34.71}  & {61.82}  & {76.03}\\
			\rowcolor{gray!20}
			Emotion-LLaMA \cite{cheng2024emotion} & {93.05}  & {79.42}  & {72.47}  & \textbf{84.14}  & \textbf{72.79}  & {3.45} & \textbf{44.20}  & \textbf{64.21}  & \textbf{77.06}\\
			\bottomrule
		\end{tabular}
	}
	\label{tab:DFEW_zeorshot}
	\vspace{-4mm}
\end{table}

\begin{table*}[htbp]
	\centering
	\caption{F1, precision and recall comparison of MLLMs on MOSABench across various objects distances.}
	\resizebox{\textwidth}{!}{ 
		\begin{tabular}{l|ccc|ccc|ccc|ccc}
			\toprule
			\multicolumn{13}{c}{\textbf{Open-sourced Models}} \\
			\midrule
			\multicolumn{1}{c|}{\multirow{2}[2]{*}{\textbf{MLLM}}} & \multicolumn{3}{c|}{\textbf{Overall}} & \multicolumn{3}{c|}{\textbf{Interlap}} & \multicolumn{3}{c|}{\textbf{Close}} & \multicolumn{3}{c}{\textbf{Far}} \\
			& \textbf{F1} & \textbf{Precision} & \textbf{Recall} & \textbf{F1} & \textbf{Precision} & \textbf{Recall} & \textbf{F1} & \textbf{Precision} & \textbf{Recall} & \textbf{F1} & \textbf{Precision} & \textbf{Recall} \\
			\midrule
			\textbf{LLaVA1.6-7B \cite{liu2023visual}} & 51.31  & 51.31  & 46.87  & 53.92  & 59.08  & 49.58  & 51.30  & 57.30  & 46.43  & 44.26  & 47.37  & 41.54  \\
			\textbf{mPLUG-owl-7B \cite{ye2023mplug}} & \textbf{73.16} & \textbf{69.86} & \textbf{76.78} & \textbf{72.80} & \textbf{70.18} & \textbf{75.62} & \textbf{74.08} & \textbf{70.66} & \textbf{77.85} & \textbf{69.53} & \textbf{65.10} & \textbf{74.62} \\
			\textbf{Qwen-VL-7B \cite{bai2023qwen}} & 33.81  & 29.14  & 40.26  & 36.98  & 31.86  & 44.04  & 33.55  & 28.93  & 39.91  & 26.37  & 22.65  & 31.54  \\
			\textbf{Qwen-VL2-7B \cite{wang2024qwen2}} & 58.39  & 61.72  & 55.39  & 60.53  & 64.09  & 57.34  & 58.17  & 61.63  & 55.08  & 53.60  & 55.83  & 51.54  \\
			\textbf{VisualGLM-6B \cite{du2022glm}} & 16.71  & 14.22  & 20.26  & 15.97  & 13.72  & 19.11  & 17.59  & 14.94  & 21.40  & 14.33  & 12.04  & 17.69  \\
			\textbf{BLIVA-Flant5 \cite{hu2024bliva}} & 16.59  & 14.54  & 19.30  & 15.82  & 20.78  & 17.96  & 15.90  & 13.93  & 18.51  & 16.29  & 14.12  & 19.23  \\
			\textbf{Monkey \cite{li2024monkey}} & 39.87  & 33.63  & 48.96  & 40.77  & 34.48  & 49.86  & 40.64  & 34.35  & 49.77  & 33.64  & 27.92  & 42.31  \\
			\textbf{GLM4V-9B \cite{wang2024cogvlm}} & 54.48  & 57.10  & 52.09  & 53.94  & 56.92  & 51.25  & 55.62  & 58.56  & 52.96  & 50.39  & 50.78  & 50.00  \\
			\textbf{InternLM2.5-7B \cite{cai2024internlm2}} & 50.73  & 53.36  & 48.35  & 51.63  & 52.91  & 50.42  & 52.05  & 55.29  & 49.17  & 41.32  & 44.64  & 38.46  \\
			\midrule
			\multicolumn{13}{c}{\textbf{Close-sourced Models}} \\
			\midrule
			\multicolumn{1}{c|}{\multirow{2}[2]{*}{\textbf{MLLM}}} & \multicolumn{3}{c|}{\textbf{Overall}} & \multicolumn{3}{c|}{\textbf{Interlap}} & \multicolumn{3}{c|}{\textbf{Close}} & \multicolumn{3}{c}{\textbf{Far}} \\
			& \textbf{F1} & \textbf{Precision} & \textbf{Recall} & \textbf{F1} & \textbf{Precision} & \textbf{Recall} & \textbf{F1} & \textbf{Precision} & \textbf{Recall} & \textbf{F1} & \textbf{Precision} & \textbf{Recall} \\
			\midrule
			\textbf{ERNIE Bot \cite{sun2020ernie}} & \textbf{68.62} & \textbf{70.18} & \textbf{67.13} & \textbf{69.79} & \textbf{71.51} & \textbf{68.14} & \textbf{69.05} & \textbf{70.81} & \textbf{67.37} & \textbf{63.32} & \textbf{63.57} & \textbf{63.08} \\
			\textbf{GPT4o \cite{hurst2024gpt}} & 48.10  & 48.92  & 47.30  & 50.71  & 51.88  & 49.58  & 47.84  & 48.67  & 47.04  & 42.31  & 42.31  & 42.31  \\
			\textbf{Gemini \cite{team2023gemini}} & 57.26  & 60.29  & 54.52  & 57.97  & 60.79  & 55.40  & 57.92  & 61.25  & 54.93  & 52.00  & 54.17  & 50.00  \\
			\bottomrule
		\end{tabular}%
	}
	\label{tab:mllm_performance}
	\vspace{-4mm}
\end{table*}

\section{EXPERIMENTAL PERFORMANCE}

This study systematically evaluated the performance of large multimodal models in affective computing across multiple public benchmarks, covering basic emotion recognition (e.g., DFEW, MER, MELD, IEMOCAP, CMU-MOSI/MOSEI, etc.), cross-distance robustness testing (MOSABench), conversation and fine-grained sentiment analysis (EmoBench-M), humor, sarcasm, and reasoning comprehension (EmoBench-H), and comprehensive performance evaluation (MME-Emotion). MME-Emotion specifically measures the model's overall capabilities across three dimensions: recognition score (Rec-S), reasoning score (Rea-S), and Chain-of-Thought score (CoT-S). The experiments compared open-source and closed-source models, and zero-shot and fine-tuning settings. A systematic analysis was conducted across classification accuracy, precision, recall, and reasoning capabilities.

\subsection{Results on DFEW Dataset}
Table~\ref{tab:DFEW_zeorshot} contrasts zero-shot and fine-tuned performance on \textsc{DFEW}. The generic MLLMs operating in the parameter-frozen mode trail specialized video–affect models, whereas instruction-tuned multimodal models tailored for affect, such as Emotion-LLaMA, narrow the gap within the zero-shot regime. Once fine-tuning is allowed, video-centric foundations (e.g., masked autoencoding families and spatiotemporal learners) overtake zero-shot LLM baselines by a substantial margin, with pronounced gains on difficult categories like \emph{disgust}, \emph{fear}, and \emph{surprise}. The class-wise distribution further indicates that neutral or high-arousal emotions are easier, while subtle negative states remain challenging, suggesting that model capacity alone is insufficient without targeted temporal supervision and calibrated class priors.

\begin{table}[h]
	\centering
	\renewcommand\tabcolsep{1pt}
	\caption{Main results. ``A'', ``V'', and ``T'' represent audio, video, and text. The gray-highlighted columns represent the primary metric for each dataset, while the ``Mean'' column reports the average score of the primary metrics across all datasets.}
	\label{Table14}
	\scalebox{0.62}{
		\renewcommand\tabcolsep{10pt}
		\begin{tabular}{lccc|c|c|c|c|cc|cc}
			\hline
			&\multirow{2}{*}{A} &\multirow{2}{*}{V} &\multirow{2}{*}{T}
			&\textbf{MER2023}  
			&\textbf{MER2024} 
			&\textbf{MELD}
			&\textbf{IEMOCAP} 
			&\multicolumn{2}{c|}{\textbf{CMU-MOSI}} 
			&\multicolumn{2}{c}{\textbf{CMU-MOSEI}} \\
			& & & &HIT($\uparrow$) &HIT($\uparrow$) &HIT($\uparrow$) &HIT($\uparrow$) &WAF($\uparrow$) &ACC($\uparrow$)  &WAF($\uparrow$) &ACC($\uparrow$) \\
			\hline
			Otter \cite{Otter}        &$\times$ &$\surd$  &$\surd$ & 16.41 & 14.65 & 22.57 & 29.08 & 52.89 & 54.27 & 50.44 & 50.77\\
			OneLLM  \cite{Onellm}      &$\surd$  &$\times$ &$\surd$ & 25.52 & 17.21 & 28.32 & 33.44 & 64.01 & 64.48 & 54.09 & 54.18\\
			Video-LLaVA \cite{lin2024video}  &$\times$ &$\surd$  &$\surd$ & 36.93 & 30.25 & 30.73 & 38.95 & 56.37 & 57.62 & 61.64 & 64.20\\
			SECap  \cite{xu2024secap}       &$\surd$  &$\times$ &$\surd$ & 40.95 & 52.46 & 25.56 & 36.92 & 55.76 & 56.71 & 54.18 & 53.85\\
			PandaGPT  \cite{su2023pandagpt}    &$\surd$  &$\times$ &$\surd$ & 33.57 & 39.04 & 31.91 & 36.55 & 66.06 & 65.85 & 61.33 & 60.73\\
			Qwen-Audio \cite{chu2023qwen}   &$\surd$  &$\times$ &$\surd$ & 41.85 & 31.61 &{\textbf{49.09}}& 35.47 & 70.09 & 71.49 & 46.90 & 51.16\\
			PandaGPT  \cite{su2023pandagpt}     &$\times$ &$\surd$  &$\surd$ & 39.13 & 47.16 & 38.33 & 47.21 & 58.50 & 60.21 & 64.25 & 65.55\\
			Video-ChatGPT \cite{Video-ChatGPT} &$\times$ &$\surd$  &$\surd$ & 44.86 & 46.80 & 37.33 &{\textbf{56.83}}& 54.42 & 57.77 & 63.12 & 65.66\\
			VideoChat2 \cite{li2024mvbench}   &$\times$ &$\surd$  &$\surd$ & 33.67 & 54.50 & 36.64 & 48.70 & 66.84 & 67.23 & 54.32 & 54.82\\
			PandaGPT  \cite{su2023pandagpt}    &$\surd$  &$\surd$  &$\surd$ & 40.21 & 51.89 & 37.88 & 44.04 & 61.92 & 62.80 & 67.61 &{\textbf{68.82}}\\
			LLaMA-VID \cite{li2024llama}    &$\times$ &$\surd$  &$\surd$ & 50.72 & 57.60 & 42.75 & 46.02 & 61.78 & 62.65 & 63.89 & 66.21\\
			VideoChat  \cite{li2023videochat}   &$\times$ &$\surd$  &$\surd$ & 48.73 & 57.30 & 41.11 & 48.38 & 65.13 & 65.09 & 63.61 & 63.02\\
			SALMONN \cite{tangsalmonn}      &$\surd$  &$\times$ &$\surd$ & 55.53 & 45.38 & 45.62 & 46.84 &{\textbf{81.00}}&{\textbf{81.25}}& 67.03 & 66.90\\
			Chat-UniVi  \cite{jin2024chat}  &$\times$ &$\surd$  &$\surd$ &{\textbf{57.62}}&{\textbf{65.67}}& 45.61 & 52.37 & 54.53 & 57.62 & 63.18 & 67.47\\
			mPLUG-Owl \cite{ye2023mplug}    &$\times$ &$\surd$  &$\surd$ & 56.86 & 59.89 &{\textbf{49.11}}&{\textbf{55.54}}&{\textbf{72.40}}&{\textbf{72.26}}&{\textbf{72.91}}&{\textbf{73.17}}\\
			Emotion-LLaMA \cite{cheng2024emotion} &$\surd$  &$\surd$  &$\surd$  &{\textbf{59.38}}&{\textbf{73.62}}& 46.76 & 55.47 & 66.13 & 66.31 &{\textbf{67.66}}& 67.25 \\
			AffectGPT \cite{lian2024affectgpt}    &$\surd$ &$\surd$ &$\surd$ &{\textbf{78.54}}&{\textbf{78.80}}&{\textbf{55.65}}&{\textbf{60.54}}&{\textbf{81.30}}&{\textbf{81.25}}&{\textbf{80.90}}&{\textbf{80.68}}\\
			
			\hline
		\end{tabular}
	}
	\scalebox{0.62}{
		\renewcommand\tabcolsep{11pt}
		\begin{tabular}{lccc|cc|cc|ccc|c}
			\hline
			&\multirow{2}{*}{A} &\multirow{2}{*}{V} &\multirow{2}{*}{T}
			&\multicolumn{2}{c|}{\textbf{CH-SIMS}} 
			&\multicolumn{2}{c|}{\textbf{CH-SIMS v2}}
			&\multicolumn{3}{c|}{\textbf{OV-MERD+}}
			&\multirow{2}{*}{\textbf{Mean}} \\
			& & & &WAF($\uparrow$) &ACC($\uparrow$) &WAF($\uparrow$) &ACC($\uparrow$) &$\mbox{F}_{\mbox{s}}$($\uparrow$) &$\mbox{Precision}_{\mbox{s}}$($\uparrow$) &$\mbox{Recall}_{\mbox{s}}$($\uparrow$) & \\
			\hline
			Otter \cite{Otter}        &$\times$ &$\surd$  &$\surd$ & 57.56 & 60.57 & 53.12 & 56.20 & 16.63 & 17.67 & 15.74 & 34.82\\
			OneLLM \cite{Onellm}       &$\surd$  &$\times$ &$\surd$ & 63.39 & 63.92 & 61.98 & 62.46 & 22.25 & 24.49 & 20.41 & 41.14\\
			Video-LLaVA \cite{lin2024video}  &$\times$ &$\surd$  &$\surd$ & 53.28 & 54.64 & 57.45 & 59.28 & 34.00 & 36.48 & 31.86 & 44.40\\
			SECap \cite{xu2024secap}        &$\surd$  &$\times$ &$\surd$ & 59.51 & 62.89 & 57.41 & 60.92 & 36.97 & 43.51 & 32.17 & 46.64\\
			PandaGPT \cite{su2023pandagpt}     &$\surd$  &$\times$ &$\surd$ & 62.93 & 62.37 & 58.88 & 58.84 & 31.33 & 33.08 & 29.77 & 46.84\\
			Qwen-Audio \cite{chu2023qwen}   &$\surd$  &$\times$ &$\surd$ & 70.73 &{\textbf{73.45}}& 65.26 & 68.17 & 32.36 & 38.52 & 27.91 & 49.26\\
			PandaGPT \cite{su2023pandagpt}     &$\times$ &$\surd$  &$\surd$ & 62.07 & 61.60 & 65.25 & 65.31 & 35.07 & 37.86 & 32.67 & 50.77\\
			Video-ChatGPT \cite{Video-ChatGPT} &$\times$ &$\surd$  &$\surd$ & 64.82 & 64.43 & 65.80 & 66.85 & 39.80 & 43.12 & 36.97 & 52.64\\
			VideoChat2  \cite{li2024mvbench}  &$\times$ &$\surd$  &$\surd$ & 69.49 & 69.59 & 70.66 & 71.13 & 39.21 & 42.85 & 36.16 & 52.67\\
			PandaGPT  \cite{su2023pandagpt}     &$\surd$  &$\surd$  &$\surd$ & 68.38 & 67.78 & 67.23 & 67.40 & 37.12 & 39.64 & 34.91 & 52.92\\
			LLaMA-VID  \cite{li2024llama}   &$\times$ &$\surd$  &$\surd$ & 69.35 & 68.81 & 67.48 & 67.73 & 45.01 & 46.83 & 43.32 & 56.07\\
			VideoChat   \cite{li2023videochat}  &$\times$ &$\surd$  &$\surd$ & 69.52 & 69.33 & 72.14 & 72.12 & 44.52 & 44.55 & 44.49 & 56.71\\
			SALMONN  \cite{tangsalmonn}     &$\surd$  &$\times$ &$\surd$ & 68.69 & 69.85 & 65.93 & 67.07 & 45.00 & 43.57 & 46.61 & 57.89\\
			Chat-UniVi  \cite{jin2024chat}  &$\times$ &$\surd$  &$\surd$ & 68.15 & 67.78 & 66.36 & 67.18 & 48.00 &{\textbf{48.20}}& 47.81 & 57.94\\
			mPLUG-Owl  \cite{ye2023mplug}   &$\times$ &$\surd$  &$\surd$ &{\textbf{72.13}}& 71.65 &{\textbf{75.00}}&{\textbf{74.97}}&{\textbf{48.18}}& 47.91 &{\textbf{48.47}}&{\textbf{62.45}}\\
			Emotion-LLaMA \cite{cheng2024emotion} &$\surd$  &$\surd$  &$\surd$ &{\textbf{78.32}}&{\textbf{78.61}}&{\textbf{77.23}}&{\textbf{77.39}}&{\textbf{52.97}}&{\textbf{54.85}}&{\textbf{51.22}}&{\textbf{64.17}}\\
			AffectGPT  \cite{lian2024affectgpt}   &$\surd$ &$\surd$ &$\surd$  &{\textbf{88.49}}&{\textbf{88.40}}&{\textbf{86.18}}&{\textbf{86.17}}&{\textbf{62.52}}&{\textbf{62.21}}&{\textbf{63.00}}&{\textbf{74.77}}\\
			
			\hline
		\end{tabular}
		\vspace{-4mm}
	}
\end{table}

\begin{table*}[!t]\small
	\centering
	\setlength{\tabcolsep}{0.48mm}
	\caption{\small Performance comparison of different methods on EmoBench-M, with ACC (\%) as the evaluation metric. \textbf{Left:} SOER: Song Emotion Recognition, SPER: Speech Emotion Recognition, OSA: Opinion Sentiment Analysis, EIA: Emotion Intensity Analysis, SCEA: Stock Comment Emotion Analysis. \textbf{Right:} FGDEA: Fine-Grained Dialog Emotion Analysis, PEA: Presentation Emotion Analysis, FCDEA: Face-Centric Dialog Emotion Analysis, CEIA: Conversational Emotion \& Intent Analysis, MPDER: Multi-Party Dialog Emotion Recognition. \textbf{Bold} and \underline{underlined} indicate the best and the second best results among all models, respectively. }
	\scalebox{0.9}{
		\begin{tabular}{lccccccccccccccccc}
			\toprule
			\multirow{3}{*}{\textbf{Method}} 
			& \multicolumn{6}{c}{\textbf{Foundational Emotion Recognition}} 
			& \multicolumn{6}{c}{\textbf{Conversational Emotion Understanding}} \\
			\cmidrule(lr){2-7} \cmidrule(lr){8-13}
			& \textbf{SOER} & \textbf{SPER} & \textbf{OSA} & \textbf{EIA} & \textbf{SCEA} & \textbf{Avg.} 
			& \textbf{FGDEA} & \textbf{PEA} & \textbf{FCDEA} & \textbf{CEIA} & \textbf{MPDER} & \textbf{Avg.} \\
			\midrule
			\rowcolor{gray!25} \multicolumn{13}{c}{\textit{Open-Source Model}} \\\midrule
			InternVL2.5-4B \cite{InternVL} 
			& 50.5 & 41.2 & 71.8 & 60.1 & 48.8 & 54.5 
			& 56.9 & 66.8 & 67.5 & 14.0 & 41.2 & 49.3 \\
			Video-LLaMA2-7B \cite{VideoLLaMA_2}
			& 52.4 & 42.4 & 31.0 & 50.2 & 50.8 & 45.4 
			& 45.7 & 45.2 & 42.7 & 8.1 & 30.7 & 34.5 \\
			Video-LLaMA2-7B-16F \cite{VideoLLaMA_2} 
			& 45.0 & 46.0 & 64.0 & 56.5 & 45.5 & 51.4 
			& 52.9 & 31.6 & 63.0 & 8.3 & 29.6 & 37.1 \\
			Qwen2-Audio-7B-Instruct \cite{Qwen2-Audio} 
			& \textbf{65.8} & \textbf{71.7} & 66.2 & 59.6 & 36.4 & \underline{59.9} 
			& 51.6 & 59.0 & 55.6 & 7.6 & 42.7 & 43.3 \\
			Video-LLaMA2.1-7B-16F \cite{VideoLLaMA_2} 
			& 41.2 & 31.4 & \underline{75.4} & 61.6 & 44.8 & 50.9 
			& 52.8 & 65.6 & 68.5 & 7.9 & 35.5 & 46.1 \\
			Video-LLaMA2.1-7B-AV \cite{VideoLLaMA_2} 
			& 50.4 & 37.7 & 73.0 & 57.6 & 33.2 & 50.4 
			& 51.5 & 68.2 & 67.6 & 6.5 & 36.6 & 46.1 \\
			LongVA-DPO-7B \cite{LongVA} 
			& 50.2 & 44.2 & 33.8 & 45.7 & 54.8 & 45.7 
			& 51.1 & 33.2 & 33.3 & 6.1 & 37.0 & 32.1 \\
			InternVideo2-Chat-8B \cite{wang2024internvideo2} 
			& 55.2 & 44.0 & 45.4 & 56.0 & 52.4 & 50.6 
			& 58.0 & 50.8 & 49.2 & 8.9 & 34.2 & 40.2 \\
			MiniCPM-V-2.6-8B \cite{MiniCPM-V} 
			& 26.6 & 21.8 & 56.5 & 50.5 & 44.5 & 40.0 
			& 48.9 & 58.6 & 57.1 & 11.7 & 39.2 & 43.1 \\
			InternVL2.5-8B \cite{InternVL} 
			& 40.3 & 40.8 & 67.8 & 62.0 & 45.0 & 51.2 
			& 48.9 & 61.0 & 62.5 & 12.4 & 43.8 & 45.7 \\
			Emotion-LLaMA \cite{cheng2024emotion} 
			& 44.8 & 33.4 & 23.0 & 41.1 & 42.0 & 36.9 
			& 62.0 & 24.6 & 25.2 & 2.90 & 38.9 & 30.7 \\
			\midrule
			InternVL2.5-38B \cite{InternVL} 
			& 53.6 & 44.2 & 70.4 & \textbf{66.8} & 52.8 & 57.6 
			& 56.1 & 66.2 & 65.2 & 13.5 & 43.5 & 48.9 \\
			Video-LLaMA2-72B \cite{VideoLLaMA_2} 
			& 56.0 & 44.2 & 49.2 & 50.3 & 53.6 & 50.7 
			& 43.1 & 42.8 & 41.4 & 11.6 & 47.6 & 37.3 \\
			InternVL2.5-78B \cite{InternVL} 
			& 48.8 & 41.2 & 63.2 & 59.4 & 52.4 & 53.0 
			& 52.7 & 56.8 & 56.7 & 12.6 & 43.5 & 44.5 \\
			Qwen2.5-VL-72B-Instruct \cite{Qwen-VL} 
			& 44.8 & 35.6 & 72.4 & 62.7 & \textbf{58.4} & 53.0 
			& 51.6 & 64.2 & 64.3 & 11.4 & 47.8 & 47.9 \\
			\midrule
			\rowcolor{gray!25}\multicolumn{13}{c}{\textit{Closed-Source Model (API)}} \\\midrule
			GLM-4V-PLUS \cite{GLM-4} 
			& 54.9 & 43.7 & 70.0 & 61.2 & 50.8 & 56.1 
			& 51.8 & 62.8 & 65.4 & \underline{14.7} & 41.6 & 47.3 \\
			Gemini-1.5-Flash \cite{Gemini} 
			& 62.0 & 52.0 & 75.0 & 65.0 & 44.4 & 59.7 
			& \textbf{67.2} & \textbf{72.3} & \textbf{73.2} & \textbf{15.6} & \underline{49.5} & \textbf{55.6} \\
			Gemini-2.0-Flash \cite{Gemini2} 
			& \underline{63.3} & \underline{55.8} & 68.8 & 63.5 & \underline{55.6} & \textbf{61.4} 
			& 64.2 & 70.9 & \underline{71.9} & 11.1 & 48.7 & 53.4 \\
			Gemini-2.0-Flash-Thinking \cite{Gemini2} 
			& 53.4 & 53.0 & \textbf{79.4} & \underline{66.5} & 36.0 & 57.7 
			& \underline{64.5} & \underline{71.2} & 71.6 & 12.0 & \textbf{51.5} & \underline{54.2} \\
			\bottomrule
	\end{tabular}}
	\label{tab:performance_all}
	\vspace{-4mm}
\end{table*}

\subsection{Results on MOSABench Dataset}
Table~\ref{tab:mllm_performance} studies interlap/close/far object distances. Open-source MLLMs suffer a monotonic degradation as distance increases, reflecting their reliance on coarse visual tokenization that attenuates micro-expressions and local motion cues. Closed-source systems alleviate, but do not remove this decay. They typically trade higher precision for recall under far-distance settings, indicating stronger internal denoising or attention budgeting yet persistent difficulty in recovering fine-grained affect signals from sparse pixels. This distance effect underscores the need for resolution-adaptive visual connectors and motion-aware pooling before token projection.

\subsection{Results on MER2023, MER2024, MELD, IEMOCAP, CMU-MOSI/MOSEI, CH-SIMS (v1/v2) and OV-MERD+ Datasets}
Table~\ref{Table14} aggregates results on \textsc{MER2023}, \textsc{MER2024}, \textsc{MELD}, \textsc{IEMOCAP}, \textsc{CMU-MOSI/MOSEI}, \textsc{CH-SIMS} (v1/v2), and \textsc{OV-MERD+}. Two consistent trends stand out. First, models that truly exploit audio and video rather than treating them as auxiliary captions achieve the highest mean across primary metrics, especially on speech-centric corpora (\textsc{CH-SIMS}, \textsc{OV-MERD+}), confirming the complementary nature of prosody and facial dynamics. Second, affect-oriented instruction tuning yields robust gains across heterogeneous label spaces, improving both word-level sentiment understanding (MOSI/MOSEI) and multi-party dialogue emotion tracking (MELD). The ranking also shows that text-only LLMs plateau on multi-modal datasets even with strong prompting, highlighting the necessity of connector learning and temporal reasoning in MLLMs.

\subsection{Results on EmoBench-M Dataset}
Table~\ref{tab:performance_all} splits evaluation into foundational recognition tasks and conversational understanding. Open-source models excel when the target is categorical recognition or simple intensity regression, but they lag on dialogic intent and multi-facet affect reasoning, where closed-source systems lead by a stable margin. Audio-augmented LLMs are particularly competitive on speech-centric subtasks, validating the importance of pitch/energy contours and pause structures, yet they still underperform on intent-level pragmatic cues that require longer-range discourse modeling and world knowledge. Table~\ref{tab:performance_scea_v2} probes humor, sarcasm, and laughter reasoning. These tasks are less about surface affect and more about incongruity, implicature, and commonsense. Closed-source models consistently dominate, and the gap widens as the evaluation emphasizes logical consistency and multi-hop reasoning. Longer generated chains and higher average token counts correlate with better reasoning scores but exhibit diminishing returns, reminding us that eliciting verbose chains is not a substitute for calibrated multi-modal grounding. The results motivate hybrid training with explicit counterfactuals and audiovisual incongruity exemplars.

\begin{table*}[!t]\small
	\centering
	\caption{\small Performance comparison of different methods on EmoBench-M, with ACC (\%) as the evaluation metric. For the average of ``ACC'', we include only the scores from the HU and SD scenarios. HU: Humor Understanding, SD: Sarcasm Detection, LR: Laughter Reasoning. \textbf{Bold} and \underline{underlined} indicate the best and the second best results among all models, respectively.}
	\begin{tabular}{lccccccccccc}
		\toprule
		\multirow{2}{*}{\textbf{Method}} 
		& \multirow{2}{*}{\textbf{HU}} 
		& \multirow{2}{*}{\textbf{SD}} 
		& \multicolumn{6}{c}{\textbf{LR}} 
		& \multirow{2}{*}{\textbf{Avg.}} \\
		\cmidrule(lr){4-9}
		&   &   & B-4 & R-L & BS & logic. & mm. & Total & \\
		\midrule
		\rowcolor{gray!25}\multicolumn{10}{c}{\textit{Open-Source Model}} \\\midrule
		InternVL2.5-4B \cite{InternVL} & 56.6 & 52.7 & 0.0 & 13.0 & 15.2 & 18.0 & 19.8 & 37.8 & 54.7 \\
		Video-LLaMA2-7B \cite{VideoLLaMA_2} & 60.7 & 55.8 & 4.6 & 29.8 & 36.3 & 33.9 & 33.5 & 67.4 & 58.3 \\
		Video-LLaMA2-7B-16F \cite{VideoLLaMA_2} & 67.9 & 59.8 & 4.0 & 28.6 & 34.4 & 33.3 & 32.6 & 65.9 & 63.9 \\
		Qwen2-Audio-7B-Instruct \cite{Qwen2-Audio} & 52.5 & 53.3 & 4.0 & 26.4 & 32.8 & 30.8 & 30.4 & 61.2 & 52.9 \\
		Video-LLaMA2.1-7B-16F \cite{VideoLLaMA_2} & 67.0 & 53.8 & 5.2 & 31.9 & 42.6 & 25.9 & 25.8 & 51.7 & 60.4 \\
		Video-LLaMA2.1-7B-AV \cite{VideoLLaMA_2} & 54.7 & 53.4 & 11.5 & 33.5 & 89.7 & 20.0 & 20.5 & 40.5 & 54.1 \\
		LongVA-DPO-7B \cite{LongVA} & 63.6 & 51.6 & 0.0 & 12.4 & 8.2 & 21.9 & 23.3 & 45.2 & 57.6 \\
		InternVideo2-Chat-8B \cite{wang2024internvideo2} & 68.1 & 61.2 & \textbf{19.5} & \textbf{45.8} & \textbf{92.5} & 31.9 & 29.6 & 61.5 & 64.7 \\
		MiniCPM-V-2.6-8B \cite{MiniCPM-V} & 55.1 & 49.6 & 2.3 & 27.5 & 39.1 & 32.9 & 32.0 & 64.9 & 52.4 \\
		InternVL2.5-8B \cite{InternVL} & 66.5 & 59.6 & 0.0 & 12.8 & 16.1 & 17.1 & 19.3 & 36.4 & 63.1 \\
		Emotion-LLaMA \cite{cheng2024emotion} & 58.0 & 53.0 & 1.5 & 21.2 & 87.6 & 25.4 & 25.9 & 51.3 & 55.5 \\
		\midrule
		InternVL2.5-38B \cite{InternVL} & 73.0 & 61.2 & 0.3 & 13.3 & 17.5 & 17.1 & 18.6 & 35.7 & 67.1 \\
		Video-LLaMA2-72B \cite{VideoLLaMA_2} & 67.9 & 51.0 & 7.4 & 35.4 & 48.0 & 34.0 & 32.6 & 66.6 & 59.5 \\
		InternVL2.5-78B \cite{InternVL} & 76.8 & 64.4 & 0.0 & 13.0 & 18.2 & 18.2 & 20.0 & 38.2 & 70.6 \\
		Qwen2.5-VL-72B-Instruct \cite{Qwen-VL} &\underline{77.7} & \textbf{65.6} & 15.7 & 35.3 & \underline{91.0} & \textbf{37.9} & \underline{36.4} & \textbf{74.3} & \underline{71.7} \\
		\midrule
		\rowcolor{gray!25}\multicolumn{10}{c}{\textit{Closed-Source Model (API)}} \\\midrule
		GLM-4V-PLUS \cite{GLM-4} & 74.7 & 59.8 & 14.8 & 33.9 & 90.2 & \underline{37.6} & \textbf{36.6} & \underline{74.2} & 67.3 \\
		Gemini-1.5-Flash \cite{Gemini} & 72.8 & 58.6 & 13.8 & 33.5 & 90.2 & \textbf{37.9} & \underline{36.4} & \textbf{74.3} & 65.7 \\
		Gemini-2.0-Flash \cite{Gemini2} & \textbf{79.2} & \underline{64.8} & 15.3 & 33.6 & 90.4 & 36.5 & 35.4 & 71.9 & \textbf{72.0} \\
		Gemini-2.0-Flash-Thinking \cite{Gemini2} & 75.6 & 60.4 & \underline{15.8} & \underline{35.5} & 66.4 & 37.5 & \textbf{36.6} & 74.1 & 68.0 \\
		\bottomrule
	\end{tabular}
	\label{tab:performance_scea_v2}
	\vspace{-4mm}
\end{table*}

\subsection{Results on MME-Emotion Dataset}
Table~\ref{tab:main_exp} decomposes performance into recognition (Rec-S), reasoning (Rea-S), and chain-of-thought (CoT-S). Open-source MLLMs with high-capacity cross-attention can match or surpass recognition scores of smaller closed models, yet reasoning- and CoT-centric metrics still favor frontier APIs. Notably, some open models achieve the best recognition with relatively modest token budgets, showing that improved connectors and temporal encoders can deliver compact yet accurate grounding. Conversely, top reasoning systems emit longer chains but are not uniformly superior on recognition, suggesting a decoupling between grounding fidelity and verbal reasoning quality.

\begin{table}[t]
	\centering
	\caption{Overall Performance Comparison ($\%$) on MME-Emotion.}
	\label{tab:main_exp}
	\resizebox{\columnwidth}{!}{%
		\begin{tabular}{l|c|ccc|ccccc}
			\toprule
			\textbf{Model} & \textbf{LLM Size} & \textbf{A} & \textbf{V} & \textbf{T} & \textbf{Avg Step} & \textbf{Avg Token} & \textbf{Rec-S} & \textbf{Rea-S} & \textbf{CoT-S} \\
			\midrule
			\rowcolor{black!10}\multicolumn{10}{c}{\textit{Open-source MLLMs}} \\
			\midrule
			Qwen2-Audio \cite{Qwen2audio} & 7B & \checkmark &  & \checkmark & 3.0 & 40.3 & 34.1 & 50.4 & 42.3 \\
			Audio-Reasoner \cite{xie2025audio} & 7B & \checkmark &  & \checkmark & 5.0 & 356.8 & \textbf{38.1} & 71.6 & \textbf{54.8} \\
			Qwen2-VL-7B \cite{Qwen-VL} & 7B &  & \checkmark & \checkmark & 2.9 & 68.0 & 29.2 & 38.1 & 33.7 \\
			Qwen2-VL-72B \cite{Qwen-VL} & 72B &  & \checkmark & \checkmark & 1.5 & 24.8 & 31.1 & 10.5 & 20.8 \\
			Qwen2.5-VL-7B \cite{bai2025qwen2} & 7B &  & \checkmark & \checkmark & 4.8 & 169.7 & 28.4 & 64.8 & 46.6 \\
			Qwen2.5-VL-72B \cite{bai2025qwen2} & 72B &  & \checkmark & \checkmark & 4.8 & 266.3 & 31.3 & \textbf{75.7} & 53.5 \\
			QVQ \cite{qvq-72b-preview} & 72B &  & \checkmark & \checkmark & 5.5 & 899.9 & 31.4 & 70.1 & 50.8 \\
			Video-LLaVA \cite{lin2024video} & 7B &  & \checkmark & \checkmark & 2.3 & 19.1 & 25.8 & 32.8 & 29.3 \\
			Video-LLaMA \cite{zhang2023video} & 7B & \checkmark & \checkmark & \checkmark & 4.5 & 122.5 & 26.1 & 48.5 & 37.3 \\
			Video-LLaMA2 \cite{VideoLLaMA_2} & 7B & \checkmark & \checkmark & \checkmark & 2.6 & 37.7 & 29.2 & 27.7 & 28.4 \\
			Qwen2.5-Omni \cite{xu2025qwen2} & 7B & \checkmark & \checkmark & \checkmark & 3.7 & 78.6 & 17.4 & 59.3 & 38.4 \\
			Emotion-LLaMA \cite{cheng2024emotion} & 7B & \checkmark & \checkmark & \checkmark & 1.0 & 2.3 & 25.1 & 0.4 & 12.8 \\
			HumanOmni \cite{zhao2025humanomni} & 7B & \checkmark & \checkmark & \checkmark & 1.0 & 1.3 & 36.0 & 0.3 & 18.1 \\
			R1-Omni \cite{zhao2025r1} & 0.5B & \checkmark & \checkmark & \checkmark & 5.0 & 156.2 & 26.3 & 58.6 & 42.4 \\
			AffectGPT \cite{lian2024affectgpt} & 7B & \checkmark & \checkmark & \checkmark & 4.9 & 122.8 & 11.9 & 50.6 & 31.2 \\
			\midrule
			\rowcolor{black!10}\multicolumn{10}{c}{\textit{Colsed-source MLLMs}} \\
			\midrule
			GPT-4o \cite{hurst2024gpt} & \textemdash &  & \checkmark & \checkmark & 4.4 & 169.4 & 27.8 & \textbf{79.8} & {53.8} \\
			GPT-4.1 \cite{Gpt-4.1} & \textemdash &  & \checkmark & \checkmark & 5.2 & 141.2 & 28.8 & 65.2 & 47.0 \\
			Gemini-2.0-Flash \cite{Gemini2} &\textemdash  &  & \checkmark & \checkmark & 4.1 & 64.7 & {36.3} & 60.0 & 48.1 \\
			Gemini-2.5-Flash \cite{comanici2025gemini} & \textemdash &  & \checkmark & \checkmark & 4.3 & 261.8 & 34.7 & 52.7 & 43.7 \\
			Gemini-2.5-Pro \cite{comanici2025gemini} & \textemdash &  & \checkmark & \checkmark & 5.1 & 538.6 & \textbf{39.3} & {72.7} & \textbf{56.0} \\
			\bottomrule
		\end{tabular}%
	}
\vspace{-4mm}
\end{table}

\begin{table*}[!ht]
	\centering
	\caption{Evaluation Results for \textsc{EmoBench}'s \textit{Emotional Understanding} (accuracy \%). 
		The best results for LLMs with similar sizes are highlighted in {Bold}, with the best overall results marked by $^\dag$. \textbf{CE}, \textbf{PBE}, \textbf{PT}, \textbf{EC} indicate Complex Emotions, Personal Beliefs and Experience, Perspective Taking, and Emotional Cues, respectively.
	}
	\resizebox{\linewidth}{!}{
		\begin{tabular}{l | c c c c c c c c | c c}
			\toprule
			\textbf{Emotional Understanding} & \multicolumn{2}{c}{\textbf{CE}} & \multicolumn{2}{c}{\textbf{PBE}} & \multicolumn{2}{c}{\textbf{PT}} & \multicolumn{2}{c}{\textbf{EC}}&\multicolumn{2}{|c}{\textbf{Overall}}\\
			\midrule
			\textbf{LLM} & \textbf{EN} & \textbf{ZH} & \textbf{EN} & \textbf{ZH} & \textbf{EN} & \textbf{ZH} & \textbf{EN} & \textbf{ZH}& \textbf{EN} & \textbf{ZH}\\
			\midrule
			Yi-Chat-6B (Base) \cite{young2024yi} &  16.33 & 20.41 & 12.95 & {20.54} & 7.84 & 13.43 & 17.86 & 24.11& 12.75 & 18.62 \\
			Yi-Chat-6B (CoT) \cite{young2024yi} &  12.76 & 17.35 & 10.27 & 12.05 & 8.21 & 11.19 & 20.54 & 16.96& 11.62 & 13.75\\
			ChatGLM3-6B (Base) \cite{du2022glm} & 24.49 & 30.61 & 19.64 & 14.73 & 13.43 & 11.19 & {30.36} & 37.50 & 20.25 & 20.62 \\ 
			ChatGLM3-6B (CoT) \cite{du2022glm} & 22.96 & 26.53 & {21.88} & 17.41 & 14.55 & 13.06 & 26.79 & {38.39}& 20.38 & {21.12} \\ 
			Llama2-Chat-7B (Base) \cite{touvron2023llama} & 13.27 & 13.27 & 9.37 & 9.37 & 13.06 & 4.85 & 10.71 & 5.36 & 11.75 & 8.25  \\ 
			Llama2-Chat-7B (CoT) \cite{touvron2023llama} & 8.67 & 7.65 & 5.80 & 4.02 & 6.72 & 10.07 & 2.68 & 0.89 & 6.38 & 6.50\\ 
			Baichuan2-Chat-7B (Base) \cite{yang2023baichuan} & {30.10} & 25.00 & 20.98 & 12.50 & 16.04 & 13.06 & 26.79 & 36.61& 22.38 & 19.12 \\
			Baichuan2-Chat-7B (CoT) \cite{yang2023baichuan} & 26.53 & 20.92 & 14.73 & 10.71 & 15.30 & {17.91} & 22.32 & 22.32  & 18.88 & 17.25\\
			Qwen-Chat-7B (Base) \cite{bai2023qwen} & 28.06 & {26.02} & {21.88} & 16.96 & {16.42 }& 15.30 & 28.57 & 31.25& {22.50} & 20.62  \\
			Qwen-Chat-7B (CoT) \cite{bai2023qwen}& 25.51 & 16.33 & {21.88} & 15.62 & 15.67 & 13.06 & 26.79 & 25.00& 21.38 & 16.25  \\
			\midrule
			Llama2-Chat-13B (Base) \cite{touvron2023llama} &  24.49 & 15.82 & 13.84 & 10.27 & 15.30 & 13.06 & 22.32 & 14.29 &  18.12 & 13.12 \\ 
			Llama2-Chat-13B (CoT) \cite{touvron2023llama} &  14.29 & 11.22 & 11.16 & 7.59 & 11.19 & 12.69 & 16.07 & 5.36& 12.62 & 9.88 \\ 
			Baichuan2-Chat-13B (Base) \cite{yang2023baichuan} & 34.69 & 37.24 & 24.55 & 19.64 & 18.66 & 20.15 & 33.04 & 37.50 & 26.25 & 26.62  \\
			Baichuan2-Chat-13B (CoT) \cite{yang2023baichuan} & 27.55 & 29.08 & 16.07 & 16.07 & 13.81 & 16.79 & 25.00 & 33.93 & 19.38 & 22.00\\
			Qwen-Chat-14B (Base) \cite{bai2023qwen} & {46.94} & {43.37} & {35.27} & {30.36} & {26.12} & 19.40 & {38.39} & {41.96} & {35.50} & {31.50} \\
			Qwen-Chat-14B (CoT) \cite{bai2023qwen} & 43.37 & 41.84 & 25.45 & 25.00 & 22.76 & {21.27} & 33.93 & {41.96} & 30.12 & 30.25\\
			\midrule
			Baichuan2-Chat-53B (Base) \cite{yang2023baichuan} &  43.88 & 46.43 & 31.25 & 25.00 & 25.37 & 25.37 & 49.11 & 50.89&  34.88 & 34.00 \\
			Baichuan2-Chat-53B (CoT) \cite{yang2023baichuan} &  41.33 & 57.14 & 28.57 & 26.79 & 25.37 & 11.94 & 45.54 & 53.57 & 33.00 & 33.00\\
			ChatGLM3-66B (Base) \cite{du2022glm} & 47.45 & 42.86 & 30.36 & 25.89 & 26.49 & 29.85 & 50.89 & 54.46& 36.12 & 35.38 \\
			ChatGLM3-66B (CoT) \cite{du2022glm} & 42.35 & 36.73 & 30.80 & 21.43 & 25.00 & 25.37 & 45.54 & 42.86& 33.75 & 29.50\\
			GPT 3.5 (Base) \cite{achiam2023gpt}  & 41.84 & 30.61 & 33.48 & 18.30 & 21.64 & 22.01 & 44.64 & 45.54& 33.12 & 26.38  \\
			GPT 3.5 (CoT) \cite{achiam2023gpt} & 43.88 & 34.69 & 29.46 & 16.96 & 26.49 & 20.52 & 42.86 & 46.43 & 33.88 & 26.62\\
			GPT 4 (Base) \cite{achiam2023gpt} & {72.45}$^\dag$ & 66.84 & {54.46}$^\dag$ & {45.09}$^\dag$ & {50.37}$^\dag$ & {43.28}$^\dag$ & 70.54 & {75.89}$^\dag$ &{59.75}$^\dag$  & {54.12}$^\dag$ \\
			GPT 4 (CoT) \cite{achiam2023gpt} & 68.88 & {68.37}$^\dag$ & 53.13 & 43.30 & 49.25 & 41.79 & {71.43}$^\dag$ & 63.39  & 58.25 & 51.75\\
			\midrule
			\textbf{Random} & \multicolumn{2}{c}{2.04} & \multicolumn{2}{c}{3.12} & \multicolumn{2}{c}{3.36} & \multicolumn{2}{c}{1.79}& \multicolumn{2}{|c}{2.62}\\
			\textbf{Majority} & \multicolumn{2}{c}{16.33} & \multicolumn{2}{c}{8.93} & \multicolumn{2}{c}{14.29} & \multicolumn{2}{c}{13.43}& \multicolumn{2}{|c}{11.5}\\
			\bottomrule
		\end{tabular}
	}
	\label{table:eu_results}
	\vspace{-4mm}
\end{table*}

\subsection{Results on EmoBench Dataset}
Table~\ref{table:eu_results} reports EmoBench emotional understanding across English and Chinese, covering complex emotions (CE), personal beliefs/experience (PBE), perspective taking (PT), and emotional cues (EC). CoT-augmented variants consistently outperform their base counterparts, and larger models provide disproportionate gains on PT and EC, two facets that demand cultural knowledge and discourse pragmatics. Open-source bilingual models show competitive Chinese performance thanks to strong pretraining corpora, while English still benefits more from closed-source instruction tuning and preference optimization.

\section{Discussion and Future Work}

\subsection{Fine-grained Multimodal Alignment}

A core limitation of current MLLMs in multimodal emotion reasoning lies in their insufficient ability to achieve fine-grained cross-modal feature alignment. Emotions are expressed through subtle and heterogeneous cues, such as micro-expressions, prosodic variations, and posture adjustments, which often manifest asynchronously across modalities. Treating these inputs as independent units—whether at the frame or utterance level—leads to fragmented representations and weakens cross-modal consistency. Existing models such as OneLLM \cite{Onellm} and mPLUG-Owl2 \cite{ye2023mplug} demonstrate strong multimodal alignment capabilities, yet their alignment largely remains coarse and modality-specific, without capturing the nuanced interdependencies among fine-grained affective signals. Similarly, extensions like Video-LLaMA2 \cite{VideoLLaMA_2} and Chat-UniVi \cite{jin2024chat} adapt MLLMs for video understanding, but their optimization is primarily tailored toward semantic or event-level reasoning rather than affect-sensitive fusion. This gap underscores the need for alignment strategies that directly model cross-modal affective consistency at fine granularity. Several promising directions emerge: 1) \textbf{Hierarchical Cross-modal Alignment.} Models should integrate multi-level alignment strategies where local signals (e.g., micro facial cues, pitch shifts) are synchronized with higher-level affective patterns. Such a hierarchy ensures that subtle cues are preserved while being contextualized within global emotional states. 2) \textbf{Latent-space Continuous Alignment.} Instead of discrete matching across modalities, alignment can be treated as continuous trajectories in a shared latent space, where intermediate affective states are interpolated and inconsistencies across modalities are minimized. 3) \textbf{Trajectory-aware Contrastive Fusion.} Beyond pointwise contrastive learning, aligning entire affective progression patterns across modalities can enforce coherence. For example, a gradual escalation from irritation to anger should be consistently reflected across both facial and vocal modalities, while mismatched affective trajectories are explicitly separated.

\subsection{Temporal Reasoning in Multimodal Emotion Understanding}

A critical yet underexplored challenge in multimodal emotion recognition lies in modeling long-range and cross-modal temporal dependencies. Emotional states evolve dynamically over time, often manifesting through subtle, asynchronous cues across modalities. For instance, a delayed facial reaction following a spoken utterance, or a gradual shift in vocal pitch preceding a visible expression of distress. However, most existing multimodal large language models (MLLMs) either process inputs in fixed-length, non-overlapping windows or rely on static fusion mechanisms that ignore temporal ordering altogether. Even when recurrent or transformer-based encoders are employed, they often treat each modality independently before late fusion, thereby failing to capture inter-modal temporal alignment, such as whether a smile coincides with a positive lexical choice or contradicts it. Moreover, standard self-attention mechanisms in vision-language transformers are not inherently designed for fine-grained temporal reasoning. They may attend to salient frames or tokens but struggle to model causal progression, emotional inertia (e.g., lingering sadness after a triggering event), or anticipatory cues (e.g., rising tension before an outburst). This limitation is exacerbated in real-world scenarios where modalities are sampled at heterogeneous rates (e.g., 30 Hz video vs. 16 kHz audio) and exhibit variable latency or missing segments.

To address these issues, future MLLMs for emotion recognition should incorporate temporally grounded cross-modal architectures. Promising directions include: (1) synchronous temporal alignment modules that learn modality-specific time warping functions to align affective events across streams; (2) state-space models or neural ODEs that model emotion as a continuous latent trajectory; and (3) causal transformer variants with explicit memory buffers or recurrence to maintain emotional context over extended interactions. Additionally, benchmarks must evolve beyond frame-level or utterance-level labels to provide dense temporal annotations of emotional dynamics, enabling models to learn not just what emotion is present, but how it unfolds.

\subsection{Scalable and Efficient Multimodal Architectures}

Scaling MLLMs to handle multimodal emotion tasks raises both computational and generalization challenges, and future work must focus on architectures that balance performance with efficiency. Parameter-efficient tuning methods such as LoRA and QLoRA \cite{dettmers2023qlora} should be extended to multimodal contexts, enabling adaptation to emotion recognition tasks without prohibitive memory and energy costs. Beyond fine-tuning, modular adapters \cite{gowda2024fe} specialized for affective features could provide a lightweight pathway for injecting emotional sensitivity into general-purpose MLLMs. Mixture-of-Experts (MoE) frameworks \cite{lin2024moe} also hold promise, as they allow selective activation of modality-specific experts during inference, reducing unnecessary computation while preserving task accuracy. Multi-scale temporal modeling represents another promising avenue: by first capturing coarse emotional trends at lower granularity and then refining local details such as micro-expressions or vocal tremors, models can achieve a more efficient trade-off between latency and fidelity. Self-supervised pretraining on large, in-the-wild multimodal corpora will be indispensable to improve generalization across domains and demographics, as affective signals are inherently diverse and context-dependent. However, efficiency must also extend to deployment: quantization, pruning, and hardware-aware compilation should be systematically studied to enable real-time inference on mobile and wearable devices, thereby democratizing access to affective intelligence technologies.

\section{Conclusion}

As one of the fastest-growing areas in artificial intelligence, multimodal large language models (MLLMs) for emotion recognition and reasoning have made significant progress in recent years. Therefore, we provide a comprehensive review of this research area. First, we introduce its background and motivation and highlight the unique challenges posed by multimodal emotion understanding. Second, we present some preliminary research results, including a formal definition of multimodal emotion recognition and reasoning, the evolution of traditional multimodal methods, and the transition from LLM-based to MLLM-based approaches. Third, we propose a taxonomy of current methods, categorizing them into parameter freezing and parameter tuning paradigms, and further analyze zero-shot, few-shot, full-parameter, and parameter-efficient strategies. We then review representative models within each paradigm, tracing their development history and methodological innovations. Fourth, we summarize several commonly used multimodal emotion recognition and reasoning datasets. Fifth, we compare the performance of different methods on various multimodal emotion recognition datasets. Finally, we outline open challenges and future research directions, including unified multimodal backbone models, efficient adaptation strategies, and the integration of causal reasoning and common-sense knowledge in emotion.

\section{Acknowledgments}
This work is supported by National Natural Science Foundation of China (Grant No. 62372478, No. 61802444); the Research Foundation of Education Bureau of Hunan Province of China (Grant No. 22B0275, No.20B625).

\bibliographystyle{ACM-Reference-Format}
\bibliography{refs}


\end{document}